\documentclass[preprint,12pt]{elsarticle}

\usepackage{hyperref}

\usepackage[utf8]{inputenc}
\usepackage{graphicx,amsfonts,amssymb,amsthm,stmaryrd,times}
\usepackage{amsmath}
\usepackage{graphicx,psfrag}
\usepackage[T1]{fontenc}
\usepackage[all]{xy}
\usepackage{multirow}
\usepackage{rotating}
\usepackage{xcolor}
\usepackage{dsfont}
\usepackage{wasysym}
\usepackage{xurl}
\usepackage{type1cm}
\usepackage{mathrsfs}
\usepackage{amsmath}
\usepackage{upgreek}
\usepackage{lettrine}
\theoremstyle{definition}
\newtheorem{thm}{Theorem}
\newtheorem*{thm*}{Theorem}

\newtheorem{deff}{Definition}
\newtheorem{exmp}{Example}
\newtheorem{experiment}{Experiment}
\newtheorem{rmk}{Remark}
\def\amax{\kern 0em\hbox{\rm \kern .25em\lower.1ex\hbox{\rlap{$\vee$}}\kern -.07em\lower.2ex\hbox{$\square$}\kern.25em}}
\def\amin{\kern 0em\hbox{\rm \kern .25em\lower.1ex\hbox{\rlap{$\wedge$}}\kern -.07em\lower.2ex\hbox{$\square$}\kern.25em}}

\def\boxmax{\kern 0em\hbox{\rm \kern .25em\lower.1ex\hbox{\rlap{$\vee$}}\kern -.07em\lower.2ex\hbox{$\square$}\kern.25em}}
\def\boxmin{\kern 0em\hbox{\rm \kern .25em\lower.1ex\hbox{\rlap{$\wedge$}}\kern -.07em\lower.2ex\hbox{$\square$}\kern.25em}}
\def\dualimp{\kern 0em\hbox{\rm \kern .25em\lower.1ex\hbox{\rlap{$\Rightarrow$}}\kern 0em\lower-1.2ex\hbox{$\overline{\hspace{2ex}}$}\kern.25em}}

\def\circmax{\kern 0em\hbox{\rm \kern .25em\lower.1ex\hbox{\rlap{$\vee$}}\kern -.18em\lower-.1ex\hbox{$\bigcirc$}\kern.25em}}
\def\circmin{\kern 0em\hbox{\rm \kern .25em\lower.1ex\hbox{\rlap{$\wedge$}}\kern -.18em\lower-.0ex\hbox{$\bigcirc$}\kern.25em}}

\newcommand{\vetu}{\boldsymbol{u}}
\newcommand{\vetx}{\boldsymbol{x}}

\newcommand{\sgn}{\text{sgn}}
\newcommand{\csgn}{\text{csgn}}

\newcommand{\tsgn}{\text{tsgn}}
\newcommand{\ii}{\boldsymbol{i}}
\newcommand{\jj}{\boldsymbol{j}}
\newcommand{\kk}{\boldsymbol{k}}

\newcommand{\re}[1]{\text{Re}\left\{#1\right\}}

\newcommand{\bb}{\begin{equation}}
\newcommand{\ee}{\end{equation}}
\newcommand{\hyper}[2]{{#1}_0 + {#1}_1 \ii_1 + \ldots + {#1}_{#2} \ii_{#2}}

\journal{}

\begin{document}

\begin{frontmatter}


\title{Hypercomplex-Valued Recurrent Correlation Neural Networks}



\author{Marcos Eduardo Valle and Rodolfo Anibal Lobo}

\address{\textit{Institute of Mathematics, Statistics, and Scientific Computing} \\
\textit{University of Campinas}\\
Campinas, Brazil \\
valle@ime.unicamp.br, rodolfolobo@ug.uchile.cl
}

\begin{abstract}
Recurrent correlation neural networks (RCNNs), introduced by Chiueh and Goodman as an improved version of the bipolar correlation-based Hopfield neural network, can be used to implement high-capacity associative memories. In this paper, we extend the bipolar RCNNs for processing hypercomplex-valued data. Precisely, we present the mathematical background for a broad class of hypercomplex-valued RCNNs. Then, we provide the necessary conditions which ensure that a hypercomplex-valued RCNN always settles at an equilibrium using either synchronous or asynchronous update modes. Examples with bipolar, complex, hyperbolic, quaternion, and octonion-valued RCNNs are given to illustrate the theoretical results. Finally, computational experiments confirm the potential application of hypercomplex-valued RCNNs as associative memories designed for the storage and recall of gray-scale images. 
\end{abstract}


\end{frontmatter}


\section{Introduction} \label{sec:introduction}

Hypercomplex numbers generalize the notion of complex and hyperbolic numbers as well as quaternions, tessarines, octonions, and many other high-dimensional algebras including Clifford and Cayley-Dickson algebras \cite{brown67,kantor89,delangue92,ell07,hitzer13,vaz16}. In this paper, we use hypercomplex algebras to define the broad class of hypercomplex-valued recurrent correlation neural networks.

Although research on hypercomplex-valued neural networks dates to the early 1970s with works written in Russian by Naun Aizenberg and collaborators \cite{aizenberg71}, the first significant research in this area appeared only in the late 1980s and early 1990s with the phasor neural networks \cite{noest88a,noest88b,noest88c} and the complex-valued multistate model \cite{aizenberg92}. As pointed out by Aizenberg, besides the proper treatment of phase information, complex-valued neural networks present higher functionality and better flexibility than the real-valued counterparts \cite{aizenberg11book}. In fact, complex-valued neural networks have been successfully applied for pattern recognition \cite{nakano09,aizenberg18wcci,hirose12,hirose19a} as well as time-series prediction and medical diagnosis \cite{aizenberg16,aizenberg18a}. Besides complex-valued models, the growing area of hypercomplex-valued neural networks include hyperbolic-valued and quaternion-valued neural networks. Hyperbolic-valued neural networks, which have been developed since the late 1990s \cite{h_ritter99,buchholz00,ontrup01}, exhibit high convergence rates an has an inherent ability to learn hyperbolic rotations \cite{nitta18}. Quaternion-valued neural networks have been applied, for example, for polarimetric synthetic aperture radar (PolSAR) land classification \cite{shang14,kinugawa18} and image processing \cite{minemoto17,parcollet19a}. An up-to-date survey on quaternion-valued neural networks can be found in \cite{parcollet19air}. 

At this point, we would like to recall that extensive effort has been devoted to the development of hypercomplex-valued Hopfield neural networks. As far as we know, research on hypercomplex-valued Hopfield neural networks dates to the late 1980s \cite{noest88a,noest88b}. In 1996, Jankowski et al. \cite{jankowski96} proposed a multistate complex-valued Hopfield network with Hebbian learning that corroborated to the development of many other hypercomplex-valued networks. For example, improved learning rules for multistate complex-valued Hopfield network include the projection rule proposed by Lee \cite{lee06} and the recording recipe based on the solution of a set of inequalities proposed by M\"uezzino\v{g}lu et al. \cite{muezzinoglu03}. Recent research on multistate complex-valued Hopfield neural network include iterative learning rules \cite{isokawa18} and networks with symmetric weights \cite{kobayashi17e}. Besides the complex-valued models, Isokawa et al. proposed quaternion-valued Hopfield neural networks using either Hebbian learning or projection rule \cite{isokawa07,isokawa08a,isokawa08b,isokawa13}. Isokawa and collaborators also investigated Hopfield neural networks based on commutative quaternions, also called tessarines \cite{isokawa10}. Quaternion-valued and tessarine-valued Hopfield neural networks are active areas of research. For example, the stability of Hopfield neural networks on unit quaternions have been addressed by Valle and de Castro \cite{valle18tnnls}. Twin-multistate activation function for quaternion-valued and tessarine-value Hopfield networks have been proposed and investigated by Kobayashi \cite{kobayashi17a,kobayashi18a}. Apart from quaternion and tessarine, Hopfield neural networks based on Clifford algebras \cite{vallejo08,kuroe11a,kuroe11b,kuroe13}, including hyperbolic numbers \cite{kobayashi13,kobayashi18c,kobayashi19a,kobayashi19c}, and octonions \cite{kuroe16,castro18cnmac} have been developed in the last decade. 

As the title of the manuscript suggest, this paper focuses on recurrent correlation neural networks. Recurrent correlation neural networks (RCNNs), formerly known as recurrent correlation associative memories (RCAMs), have been introduced by Chiueh and Goodman as an improved version of the Hopfield neural network \cite{chiueh91}. Precisely, in contrast to the original Hopfield neural network which has a limited storage capacity, some RCNN models can reach the storage capacity of an ideal associative memory \cite{hassoun88}. In few words, an RCNN is obtained by decomposing the famous  Hopfield network with Hebbian learning into a two layer recurrent neural network. The first layer computes the inner product (correlation) between the input and the memorized items followed by the evaluation of a non-decreasing continuous activation function. The subsequent layer yields a weighted average of the stored items. Alternatively, certain RCNNs can be viewed as kernelized versions of the Hopfield network with Hebbian learning \cite{garcia04a,garcia04b,perfetti08}. Furthermore, RCNNs are closely related to the dense associative memory model introduced by Krotov and Hopfield to establish the duality between associative memories and deep learning \cite{krotov16,demircigil17}. 

In this paper, we introduce a broad class of hypercomplex-valued RCNNs which include the complex-valued \cite{valle14nnB} and quaternion-valued \cite{valle18wcci} RCNN models as particular instances. As the bipolar RCNNs, the new hypercomplex-valued models can be viewed as improved versions of the hypercomplex-valued Hopfield neural networks. In fact, hypercomplex-valued RCNNs not only allow for the implementation of high-capacity associative memories but also circumvent the learning problems faced by some hypercomplex-valued Hopfield networks such as the hyperbolic-valued model \cite{kobayashi19a}.

The paper is organized as follows: Next section presents some basic concepts on hypercomplex numbers. A brief review on hypercomplex-valued Hopfield neural network is given in Sections \ref{sec:HHNN}. The hypercomplex-valued RCNNs are introduced in Section \ref{sec:HRCNNs}. Examples of hypercomplex-valued RCNNs and computational experiments are provided in Section \ref{sec:examples}. The paper finishes with the concluding remarks in Section \ref{sec:concluding} and an appendix with some remarks on the implementation of hypercomplex-valued RCNNs.

\section{Basic Concepts on Hypercomplex Numbers} \label{sec:hypercomplex_numbers}

A hypercomplex number over the real field is written as
\bb \label{eq:hypercomplex} p= \hyper{p}{n}, \ee 
where $n$ is a non-negative integer, $p_0, p_1, \ldots, p_n$ are real numbers, and the symbols $\ii_1, \ii_2, \ldots, \ii_n$ are the {\it hyperimaginary units} \citep{kantor89,castro20nn}. The set of all hypercomplex numbers given by \eqref{eq:hypercomplex} is denoted in this paper by $\mathbb{H}$. Examples of hypercomplex numbers include complex numbers $\mathbb{C}$, hyperbolic numbers $\mathbb{U}$, quaternions $\mathbb{Q}$, tessarines (also called commutative quaternions) $\mathbb{T}$, and octonions $\mathbb{O}$. 

We would like to point out that a real number $\alpha \in \mathbb{R}$ can be identified with the hypercomplex number $\alpha+0\ii_1+\ldots+0\ii_n \in \mathbb{H}$. Furthermore, the real-part of a hypercomplex number $p = \hyper{p}{n}$ is the real-number defined by $\re{p} := p_0$.

A hypercomplex number {\em system} is a set of hypercomplex numbers equipped with an {\em addition} and a {\em multiplication} (or \textit{product}). 
The addition of two hypercomplex numbers $p=\hyper{p}{n}$ and $q=\hyper{q}{n}$ is defined in a component-wise manner according to the expression 
\bb \label{eq:addition} p+q=(p_0+q_0)+(p_1+q_1)\ii_1+\ldots+(p_n+q_n)\ii_n. \ee 
The product between $p$ and $q$, denoted by the juxtaposition of $p$ and $q$, is defined using the distributive law and a multiplication table. Precisely, the multiplication table specifies the product of any two hyperimaginary units:
\bb \label{eq:unitproduct}
\ii_\mu\ii_\nu=a_{\mu\nu,0}+a_{\mu\nu,1}\ii_1+\ldots+a_{\mu\nu,n}\ii_n, \quad \forall \mu,\nu \in \{1,\ldots, n\}, \ee 
Then, using the distributive law, we obtain
\begin{align}
\label{eq:multiplication}
pq = \left( p_0q_0 + \sum_{\mu,\nu = 1}^n p_\mu q_{\nu} a_{\mu\nu,0}\right) &+ \left(p_0q_1+p_1q_0 + \sum_{\mu,\nu=1}^n p_\mu q_\nu a_{\mu\nu,1} \right) \ii_1 + \ldots  \nonumber \\
& + \left(p_0q_n+p_nq_0 + \sum_{\mu,\nu=1}^n p_\mu q_\nu a_{\mu\nu,n} \right) \ii_n. 
\end{align}

We would like to point out that we can identify a hypercomplex number $p=\hyper{p}{n}$ with the $(n+1)$-tuple $(p_0,p_1,\ldots,p_n)$ of real numbers. Furthermore, a hypercomplex number system can be embedded in a $(n+1)$-dimensional vector space where the vector addition and the scalar multiplication are obtained from \eqref{eq:addition} and \eqref{eq:multiplication}, respectively.  In view of this fact, we write $\mathtt{dim}(\mathbb{H})=n+1$. Furthermore, borrowing  terminology from linear algebra, we speak of a linear operator $T:\mathbb{H} \to \mathbb{H}$ if $T(\alpha p + q) = \alpha T(p) + T(q)$, for all $p,q\in \mathbb{H}$ and $\alpha \in \mathbb{R}$. 
A linear operator that is an involution and also an anti-homomorphism is called a {\em reverse-involution} \citep{ell07}. Formally, we have the following definition:

\begin{deff}[Reverse-involution \cite{castro20nn}]
An operator $\tau:\mathbb{H} \to \mathbb{H}$ is a reverse-involution if the following identities hold true for all $p,q \in \mathbb{H}$ and $\alpha \in \mathbb{R}$:
\begin{align} 
\label{eq:nu1}
& \tau\big(\tau(p)\big) = p, & {\mbox{(involution)}}\\
\label{eq:nu3}
& \tau(pq)=\tau(q)\tau(p), & \mbox{(antihomomorphism)}\\
\label{eq:nu2}
& \tau(\alpha p+q)=\alpha\tau(p)+\tau(q). & \mbox{(linear)} 
\end{align}  
\end{deff}
%
The natural conjugate of a hypercomplex number $p = \hyper{p}{n}$, denoted by $\bar{p}$, is defined by 
\bb \label{deff:conjug} \bar{p} = p_0-p_1\ii_1-\ldots-p_n\ii_n. \ee
The \textit{natural conjugation} is an example of a reverse-involution in some hypercomplex number systems such as the complex and quaternion number systems. The identity mapping $\tau(p)=p$, for all $p \in \mathbb{H}$, is also a reverse-involution whenever the multiplication is commutative. In this case, the identity is referred to as the \textit{trivial reverse-involution}. Other examples of reverse-involutions include the \textit{quaternion anti-involutions} in quaternion algebra and the \textit{Clifford conjugation} in Clifford algebras \citep{ell07,delangue92,vaz16}. 

A reverse-involution allows us to define a symmetric bilinear form $\mathcal{B}:\mathbb{H}\times \mathbb{H} \to \mathbb{R}$ by means of the equation:
\bb \label{eq:inner-product} \mathcal{B}(p,q) = \re{\tau(p)q}, \quad \forall p,q \in \mathbb{H}.\ee
Intuitively, $\mathcal{B}$ measures a relationship between $p$ and $q$ by taking into account the algebraic properties of the multiplication and the reverse-involution $\tau$. For example, the symmetric bilinear form $\mathcal{B}$ coincides with the usual inner product on complex numbers, quaternions, and octonions with the natural conjugation. The symmetric bilinear form $\mathcal{B}$ plays an important role in the definition of the hypercomplex-valued recurrent correlation neural networks introduced in this paper.

Finally, let us also recall the following class of hypercomplex-valued functions:
\begin{deff}[$\mathcal{B}$-function \cite{castro20nn}] \label{def:B-projection}
Consider a hypercomplex number system $\mathbb{H}$ equipped with a reverse-involution $\tau$ and let $\mathcal{D} \subset \mathbb{H}$, $\mathcal{S} \subset \mathbb{H}$, and $\mathcal{B}:\mathbb{H} \times \mathbb{H} \to \mathbb{R}$ be the symmetric bilinear form defined by \eqref{eq:inner-product}. A hypercomplex-valued function $\phi: \mathcal{D} \to \mathcal{S}$ is called a \textit{$\mathcal{B}$-function}\footnote{We would like to point out that $\mathcal{B}$-functions were referred to as $\mathcal{B}$-projection functions in \cite{castro20nn}. Projections, however, are usually defined as idempotent mappings. Since we do not require $\phi$ to be idempotent, we removed the term ``projection'' in Definition \ref{def:B-projection}.} if
\bb \label{eq:class} \mathcal{B}(\phi(q),q) > \mathcal{B}(s,q), \quad  \forall q \in D, \forall s \in \mathcal{S} \setminus \left\lbrace \phi(q) \right\rbrace. \ee
\end{deff}


In words, a $\mathcal{B}$-function $\phi:\mathcal{D} \to \mathcal{S}$ maps $q \in \mathcal{D}$ to a hypercomplex number $\phi(q) \in \mathcal{S}$ which is more related to $q$ than any other element $s \in \mathcal{S}$ with respect to the symmetric bilinear form $\mathcal{B}$. Examples of $\mathcal{B}$-functions include the complex-valued signum function for complex-valued Hopfield neural networks, the function that normalizes its arguments to length one on Cayley-Dickson algebras with the natural conjugation, and the split-sign functions for some real Clifford algebras with Clifford conjugation \citep{aizenberg92,jankowski96,castro20nn}.

\section{Hypercomplex-Valued Discrete-Time Hopfield Neural Networks} \label{sec:HHNN}

The famous real-valued bipolar discrete-time {\em Hopfield neural network} (HNN) is a recurrent model which can be used to implement associative memories \cite{hopfield82}. Apart from implementing associative memories, the Hopfield neural network has been applied in control \citep{gan17,song17}, computer vision and image processing \citep{wang15,jli16}, classification \citep{pajares10,zhang17}, and optimization \citep{hopfield85,serpen08,cli16}. 

Hypercomplex-valued versions of the Hopfield network have been extensively investigated in the past years using complex numbers \citep{jankowski96,lee06}, dual numbers \citep{kuroe11b}, hyperbolic numbers \citep{kobayashi13}, tessarines \citep{isokawa10,kobayashi18a}, quaternions \citep{isokawa08a,valle18tnnls}, octonions \citep{kuroe16}, and other hypercomplex number systems \citep{vallejo08,kuroe13,popa16c}. In this section, we review the broad class of hypercomplex-valued Hopfield neural networks which include many of the aforementioned models \cite{castro20nn}. 

In order to analyze the stability of a broad class of hypercomplex-valued neural networks, Castro and Valle introduced the following class of hypercomplex number systems \citep{castro20nn}: 
\begin{deff}[Real-Part Associative Hypercomplex Number Systems] \label{def1}
A hypercomplex number system equipped with a reverse-involution $\tau$ is called a \textit{real-part associative hypercomplex number system} (Re-AHN) if the following identity holds true for any three of its elements $p,q,r$:
\bb\re{(pq)r-p(qr)}=0. \label{eq:hopfieldtype}\ee
In particular, we speak of a {\it positive semi-definite (or non-negative definite)} real-part associative hypercomplex number system if the symmetric bilinear form $\mathcal{B}$ given by \eqref{eq:inner-product} satisfies
$\mathcal{B}(p,p) \geq 0$, $\forall p \in \mathbb{H}$.
\end{deff}
%
%
Complex numbers, quaternions, and octonions are examples of real-part associative hypercomplex number systems with the natural conjugation. Real-part associative hypercomplex number systems also include the tessarines, Cayley-Dickson algebras, and real Clifford algebras \cite{castro20nn}.

Let $\mathbb{H}$ be a real-part associative hypercomplex number system and $\phi:\mathcal{D} \to \mathcal{S}$ be a $\mathcal{B}$-function where $\mathcal{S}$ is a compact subset of $\mathbb{H}$. In analogy to the traditional discrete-time Hopfield network, let $w_{ij} \in \mathbb{H}$ denote the $j$th hypercomplex-valued synaptic weight of the $i$th neuron of a network with $N$ neurons. Also, let the state of the network at time $t$ be represented by a  hypercomplex-valued column-vector $\vetx(t) = [x_1(t),\ldots,x_N(t)]^T \in \mathcal{S}^N$, that is, $x_i(t) = x_{i0}(t) + x_{i1}(t)\ii_1 + \ldots + x_{in}(t) \ii_n$ corresponds to the state of the $i$th neuron at time $t$. Given an initial state (or input vector) $\vetx(0) = [x_1(0),\ldots,x_N(0)]^T \in \mathcal{S}^N$, the hypercomplex-valued Hopfield network defines recursively the sequence $\{\vetx(t)\}_{t \geq 0}$ by means of the equation
\bb \label{eq:hopfield} x_i(t+\Delta t) = \begin{cases} \phi \big( h_i(t) \big), & h_i(t) \in \mathcal{D}, \\ x_i(t), &  \mbox{otherwise}, \end{cases} \ee
where $h_i(t)=\sum_{j=1}^{N}w_{ij} x_j(t)$ is the hypercomplex-valued activation potential of the $i$th neuron at time $t$.

Like the traditional real-valued Hopfield network, the sequence produced by \eqref{eq:hopfield} is convergent in asynchronous update mode if the synaptic weights satisfy $w_{ij}=\tau(w_{ji})$ and one of the two cases below holds true \cite{castro20nn}:
\begin{enumerate}[(a)]
\item $w_{ii}=0$ for any $i \in \{1,\ldots,N\}$. 
\item $w_{ii}$ {is a nonnegative real number for any $i \in \{1,\ldots,N\}$} and $\mathbb{H}$ is a positive semi-definite real-part associative hypercomplex number system.
\end{enumerate}

The stability analysis of Hopfield networks are particularly important for the implementation of associative memories as well as for solving optimization problems \cite{hopfield82,hopfield85,hassoun95,hassoun97}. Apart from the stability analysis, applications of the Hopfield network require an appropriate rule for determining the synaptic weights. 

In an associative memory problem, we are given a set of hypercomplex-valued vectors $\mathcal{U}  = \{\vetu^1,\ldots,\vetu^P\}$ and the task is to determine the weights $w_{ij}$'s such that $\vetu^\xi$'s are stable stationary states of the network. The Hebbian learning and the projection rule are examples of recording recipes used for the storage of real and some hypercomplex-valued vectors in Hopfield neural networks \cite{hassoun97,jankowski96,lee06,isokawa13}. On the one hand, the Hebbian learning suffers from a low storage capacity due to cross-talk between the stored items \cite{hopfield82,mceliece87,isokawa13,minemoto16}. On the other hand, although the projection rule increases the storage capacity of the real, complex-valued, and quaternion-valued Hopfield network \cite{kanter87,lee06,minemoto16}, it is not straightforward for hypercomplex-valued networks such as those based on hyperbolic numbers \cite{kobayashi19a}. The recurrent correlation neural networks introduced by Chiueh and Goodman \cite{chiueh91} and further generalized to complex numbers \cite{valle14nnB} and quaternions \cite{valle18wcci} are alternative models which can be used to implement high-capacity associative memories. The next section generalizes the recurrent correlation neural networks using a broad class of hypercomplex numbers.    

\section{Hypercomplex-Valued Recurrent Correlation Neural Networks} \label{sec:HRCNNs}

Recurrent correlation neural networks (RCNNs), formerly known as recurrent correlation associative memories, have been introduced in 1991 by Chiueh and Goodman for the storage and recall of $N$-bit vectors \cite{chiueh91}. The RCNNs have been generalized for the storage and recall of complex-valued and quaternion-valued vectors  \cite{valle14nnB,valle18wcci}. In the following, we generalize further the RCNNs for the storage and recall of hypercomplex-valued vectors.

Let $\mathbb{H}$ be a hypercomplex number system, $\phi:\mathcal{D} \to \mathcal{S}$ be a $\mathcal{B}$-function, and $f:\mathbb{R} \to \mathbb{R}$ be a real-valued continuous and monotonic non-decreasing function. Also, consider a fundamental memory set $\mathcal{U} = \{\vetu^1,\ldots,\vetu^P\}$ where
\bb u_i^\xi = u_{i0}^\xi + u_{i1}^\xi \ii_1 +\ldots + u_{in}^\xi \ii_n \in \mathcal{S}, \quad \forall i = 1,\ldots,N, \forall \xi=1,\ldots,P, \ee
are hypercomplex numbers. A hypercomplex-valued RCNN (HRCNN) defines recursively the following sequence of hypercomplex-valued vectors in $\mathcal{S}^N$ for $t \geq 0$ and $i=1,\ldots,N$:
\bb \label{eq:HRCNN} x_i(t+\Delta t) = \begin{cases} \phi \big( h_i(t) \big), & h_i(t) \in \mathcal{D}, \\ x_i(t), &  \mbox{otherwise}, \end{cases} \ee
where the activation potential of the $i$th output neuron at time $t$ is given by
\bb \label{eq:activation} h_i(t)=\sum_{\xi=1}^K w_\xi(t) u_i^\xi, \quad \forall i=1,\ldots,N,\ee
with
\bb \label{eq:weights} w_\xi(t)= f\left(\sum_{i=1}^N \mathcal{B}\big(u_i^\xi,x_i(t)\big)\right), \quad \forall \xi \in 1,\ldots,K. \ee

Examples of HRCNNs include the following straightforward generalizations of the bipolar (real-valued) RCNNs of Chiueh and Goodman \cite{chiueh91}:
\begin{enumerate}
    \item The \textit{identity} HRCNN, also called correlation HRCNN, is obtained by considering in  \eqref{eq:weights} the identity function $f_e(x)=x$.
    \item The \textit{high-order} HRCNN is determined by considering the function
    \bb f_h(x) = (1+x)^q, \quad q>1.\ee
    \item The \textit{potential-function} HRCNN is obtained by considering the function 
    \bb f_p(x) = 1/(1-x)^L,  \quad L\geq 1.\ee
    \item The \textit{exponential} HRCNN, which is obtained by considering the exponential function:
    \bb \label{eq:f_e} f_e(x) = \beta e^{\alpha x}, \quad \alpha>0, \beta>0.\ee
\end{enumerate}

Let us now study the stability of HRCNNs. The following theorem shows that a HRCNN yields a convergent sequence $\{\vetx(t)\}_{t \geq 0}$ of hypercomplex-valued vectors independently of the initial state vector $\vetx(0) \in \mathcal{S}^N$.

\begin{thm} \label{thm:Convergence}
Let $\mathbb{H}$ be a hypercomplex number system, $\phi:\mathcal{D} \to \mathcal{S}$ be a $\mathcal{B}$-function where $\mathcal{S}$ is a compact set, and $f:\mathbb{R} \to \mathbb{R}$ be a real-valued continuous and monotonic non-decreasing function. Given a fundamental memory set $\mathcal{U} = \{\vetu^1,\vetu^2,\ldots,\vetu^K\} \subset \mathcal{S}^K$, the sequence $\{\vetx(t)\}_{t \geq 0}$ given by \eqref{eq:HRCNN}, \eqref{eq:activation}, and \eqref{eq:weights} is convergent for any initial state $\vetx(0) \in \mathcal{S}^N$ using either asynchronous (sequential) or synchronous update mode.
\end{thm}

\begin{rmk}
In contrast to the hypercomplex-valued Hopfield networks, the stability analysis of the hypercomplex-valued recurrent correlation neural networks does not require that $\mathbb{H}$ is a real-part associative hypercomplex-number system. As a consequence, hypercomplex-valued RCNNs can be defined in a more general class of hypercomplex number systems.   
\end{rmk}

\begin{rmk}
Theorem \ref{thm:Convergence} is derived by showing that the function 
$E:\mathcal{S}^N \to \mathbb{R}$ given by 
\bb \label{eq:energy} E(\vetx) = -\sum_{\xi=1}^K F\left(\sum_{i=1}^N \mathcal{B}(u_i^\xi,x_i) \right), \quad \forall \vetx \in \mathcal{S}^N,\ee
where $F:\mathbb{R} \to \mathbb{R}$ is a primitive of $f$, is an energy function of the HRCNN.
\end{rmk}

\begin{proof}
First of all, since $f$ is continuous, it has a continuous primitive $F:\mathbb{R} \to \mathbb{R}$. 
From the mean value theorem and the monotonicity of $f$, we have \cite{chiueh91}:
\bb \label{eq:Fineq} F(y)-F(x) \geq f(x)(y-x),\quad \forall x,y \in \mathbb{R}.\ee

Let us show that the function $E:\mathcal{S}^N \to \mathbb{R}$ given by \eqref{eq:energy} 
is an energy function of the HRCNN, that is, $E$ is a real-valued bounded function whose values decrease along non-stationary trajectories.

Let us first show that $E$ is a bounded function. Recall that the set of hypercomplex numbers $\mathbb{H}$ inherits the topology from $\mathbb{R}^{n+1}$ by identifying $p = \hyper{p}{n}$ with the $(n+1)$-tuple $(p_0,p_1,\ldots,p_n)$. A well-known generalization of the extreme value theorem from calculus for continuous functions establishes that the continuous image of a compact set is compact. Now, as a symmetric bilinear form on a finite dimensional vector space, $\mathcal{B}$ is continuous. Since a primitive $F$ of $f$ is also continuous, the function $E$ given by \eqref{eq:energy} is continuous because it is the sum and the composition of continuous functions. In addition, since $\mathcal{S}$ is compact, the Cartesian product $\mathcal{S}^N = \mathcal{S} \times \ldots \times \mathcal{S}$ is also compact. Hence,
from the extreme value theorem, we conclude that the function $E$ given by \eqref{eq:energy} is bounded. 
Let us now show that $E$ decreases along non-stationary trajectories. To simplify notation, let $\vetx \equiv \vetx(t)$ and $\vetx' \equiv \vetx(t+ \Delta t)$ for some $t \geq 0$. Furthermore, let us suppose that a set of neurons changed their state at time $t$. Formally, suppose that $x_i '\neq x_i$ if $i \in \mathcal{I}$ and $x_i' = x_i$ for all $i \not \in \mathcal{I}$, where $\mathcal{I} \subseteq \{1,\ldots,N\}$ is a non-empty set of indexes. Using \eqref{eq:Fineq}, we conclude that the variation of $E$ satisfies the following inequality: 
\begin{align}
    \Delta E &= E(\vetx')-E(\vetx) \nonumber \\
    &= - \sum_{\xi=1}^K \left[F\left(\sum_{i=1}^N \mathcal{B}(u_i^\xi,x_i')\right) - F\left(\sum_{i=1}^N \mathcal{B}(u_i^\xi,x_i)\right)  \right] \nonumber \\
    & \leq - \sum_{\xi=1}^K  f\left(\sum_{i=1}^N \mathcal{B}(u_i^\xi,x_i)\right) \left[\sum_{i=1}^N \mathcal{B}(u_i^\xi,x_i') - \sum_{i=1}^N \mathcal{B}(u_i^\xi,x_i)  \right].
\end{align}
From \eqref{eq:weights}, recalling that $\mathcal{B}$ is a symmetric bilinear form, and using \eqref{eq:activation}, we obtain
\begin{align}
\Delta E & \leq -\sum_{\xi=1}^K w_\xi(t) \sum_{i \in \mathcal{I}} \mathcal{B}(u_i^\xi,x_i'-x_i) \nonumber \\
& = - \sum_{i \in \mathcal{I}} \mathcal{B}\left(\sum_{\xi=1}^K w_\xi(t) u_i^\xi,(x_i'-x_i) \right) \nonumber \\
& = - \sum_{i \in \mathcal{I}} \mathcal{B}(h_i(t),x_i'-x_i) \nonumber \\
& = - \sum_{i \in \mathcal{I}} \left[\mathcal{B}(h_i(t),x_i') -\mathcal{B}(h_i(t),x_i) \right] = - \sum_{i \in \mathcal{I}} \Delta_i. \label{eq:DeltaE} \end{align}
Now, if the $i$th neuron changed its state at time $t$, then 
\bb x_i' = x_i(t+\Delta t) = \phi(h_i(t)) \neq x_i(t).\ee Since $\phi$ is a $\mathcal{B}$-function, we have 
\bb \Delta_i = \mathcal{B}\Big(h_i(t),\phi\big(h_i(t)\big)\Big) - \mathcal{B}\Big(h_i(t),x_i(t)\Big) > 0, \quad \forall i \in \mathcal{I}.\ee 
The latter inequality implies that $\Delta E < 0$ if at least one neuron changes its state at time $t$, which concludes the proof of the theorem.
\end{proof}

\section{Examples and Computational Experiments} \label{sec:examples}

Let us now provide some examples and perform some computational experiments with hypercomplex-valued exponential correlation neural networks (ECNN). The source-codes as well as Jupyter notebooks of the computational experiments, implemented in \texttt{Julia language}, are available at \url{https://github.com/mevalle/Hypercomplex-Valued-Recurrent-Correlation-Neural-Networks}. See the appendix for some details in our implementation of the hypercomplex-valued ECNNs.

We would like to point out that we refrained to consider other HRCNNs due to the following: First, there is a vast literature on real-valued RCNN models based on the exponential function $f_e$ \cite{chiueh91,hassoun96,hancock98,perfetti08}. The exponential, high-order, and potential-function are non-negative parametric functions continuous and strictly increasing on the variable $x$ and characterized by an exponential on their parameter \cite{valle18wcci}. Therefore, the three functions $f_e$, $f_h$, and $f_p$ belong to the same family of functions. According to Chiueh and Goodman, the bipolar (real-valued) ECNN seems to be the RCNN most amiable to VLSI implementation \cite{chiueh91}.  
In fact, the storage capacity and noise tolerance of the bipolar ECNN have been extensively investigated \cite{chiueh91}. In addition, the storage capacity of the bipolar ECNN can reach the capacity of an ideal associative memory \cite{hassoun96}. Finally, besides being closely related to support vector machines and the kernel trick \cite{perfetti08}, the bipolar ECNN has a Bayesian interpretation \cite{hancock98}.

\subsection{Bipolar RCNNs}

Let us begin by showing that the original bipolar RCNNs (formelly known as RCAMs) of Chiueh and Goodman can be view as a particular case of the broad class of hypercomplex-valued recurrent corelation neural networks introduced in this paper \cite{chiueh91}. Also, let us provide a simple example to illustrate the differences between  synchronous and asynchronous update modes and one computational experiment to validate Theorem \ref{thm:Convergence}.

Consider the real numbers $\mathbb{R}$ equipped with the trivial reverse-involution $\tau(x)=x$ for all $x \in \mathbb{R}$. In this case, the symmetric bilinear form $\mathcal{B}:\mathbb{R} \times \mathbb{R} \to \mathbb{R}$ given by \eqref{eq:inner-product} is $\mathcal{B}(x,y) = xy$. Castro and Valle showed that the function $\sgn:\mathcal{D}=\mathbb{R} \setminus \{0\} \to \mathcal{S} = \{-1,+1\}$, which yields the signal of a non-zero real-number, is a $\mathcal{B}$-function \cite{castro20nn}. In accordance with Chiueh and Goodman \cite{chiueh91}, Theorem \ref{thm:Convergence} shows that bipolar RCNNs, including the ECNN, always settle at an equilibrium, independently of the initial state and the update mode (synchronous or asynchronous).

\begin{exmp} \label{ex:Bipolar}
As an illustrative example, we synthesized a bipolar ECNN using the fundamental memory set
\bb \label{eq:Ureal} \mathcal{U} = \left\{ \vetu^1 = \begin{bmatrix} -1 \\ -1 \\ +1 \\ +1\end{bmatrix},  
\vetu^2 = \begin{bmatrix} -1 \\ +1 \\ -1 \\-1 \end{bmatrix},
\vetu^3 = \begin{bmatrix} +1 \\ +1 \\ -1 \\ -1 \end{bmatrix}
\right\},\ee
and the parameters $\alpha=1/4$ and $\beta=1$. The dynamic of synchronously and asynchronously bipolar ECNNs is depicted in the directed graphs in Figure \ref{fig:bipolarRCNN}. In these directed graph, a node corresponds to a state of the neural network (there are 16 possible states) while an edge from node $i$ to node $j$ means that we can obtain the $j$th state from the $i$th state evolving \eqref{eq:HRCNN}. Furthermore, the fundamental memories, which corresponds to the states number 4, 5, and 13, corresponds to the gray nodes in Figure \ref{fig:bipolarRCNN}. Note that, according to Theorem \ref{thm:Convergence}, both synchronous and asynchronous bipolar ECNNs always come to rest at an equilibrium.

\begin{figure}[t] \label{fig:bipolarRCNN}
    \centering
    \begin{tabular}{c}
    a) Synchronous update mode \\    \includegraphics[scale=0.8]{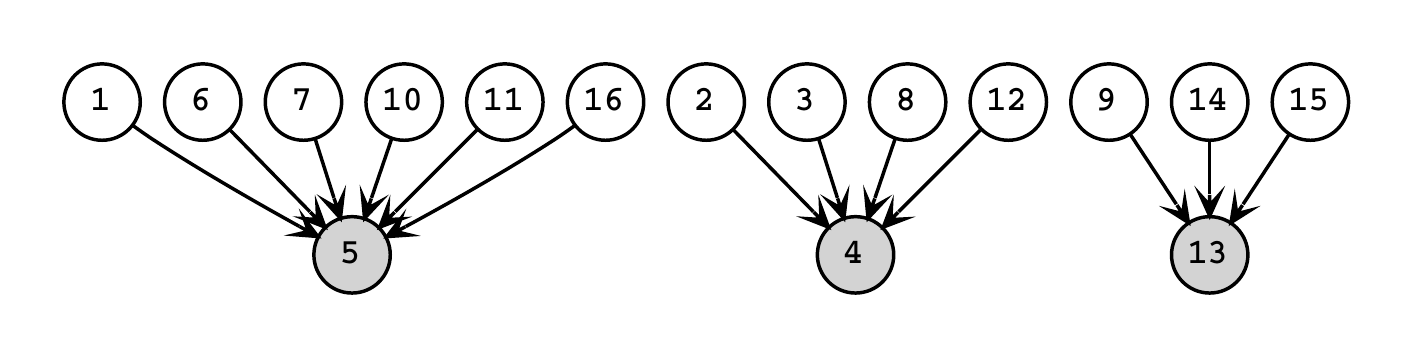} \\
    b) Asynchronous update mode \\
    \includegraphics[scale=0.8]{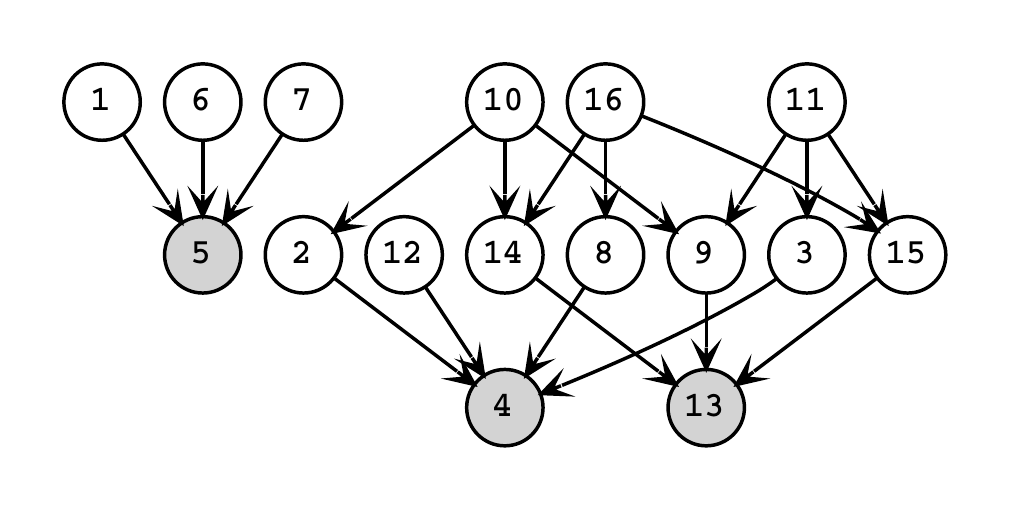}
    \end{tabular}
    \caption{Dynamic of the bipolar ECNN with $\alpha=1$ and $\beta=1$ using a) synchronous and b) asynchronous update modes.}
\end{figure}

A common problem in the design of associative memories is the creation of \textit{spurious memories} \cite{hassoun97}. Briefly, an spurious memory is a stable stationary state that does not belong to the fundamental memory set.  In this simple example, both exponential bipolar RCNNs designed for the storage of the fundamental memory set $\mathcal{U} \subset \{-1,+1\}^4$ given by \eqref{eq:Ureal} do not have any spurious memory. In fact, all the stationary states are fundamental memories. 

\end{exmp}

\begin{experiment} \label{exp:EnergyBipolar}
\begin{figure}[t] \label{fig:EnergyBipolar}
    \centering
    \includegraphics[width=0.9\columnwidth]{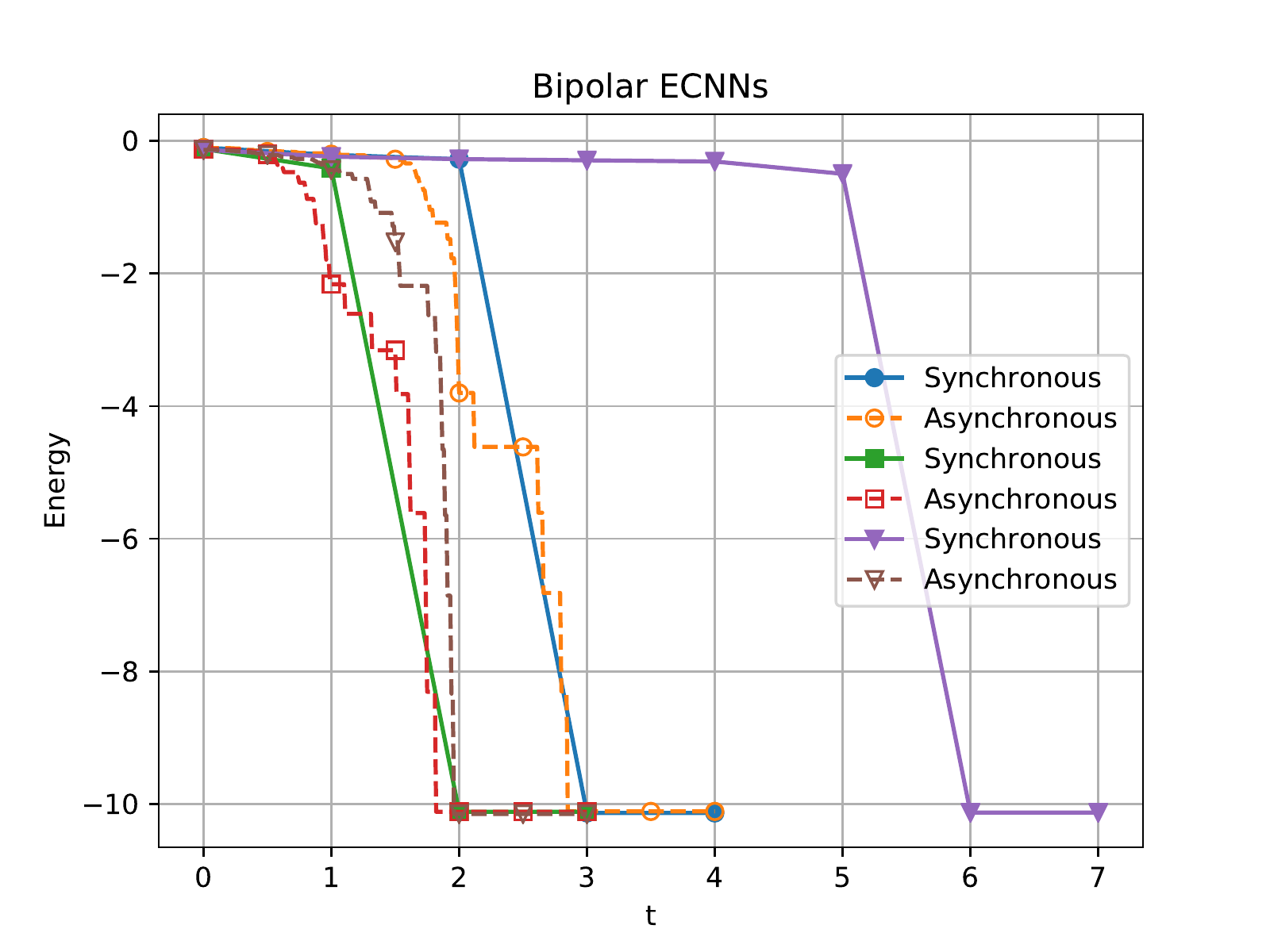}
    \caption{Evolution of the energy of the bipolar ECNN designed for the storage of $P=160$ uniformly generated fundamental memories of length $N=100$.}
\end{figure}
As a simple computational experiments, we synthesized bipolar ECNNs designed for the storage and recall of a fundamental memory set $\mathcal{U} = \{\vetu^1,\ldots,\vetu^P\} \subset \{-1,+1\}^N$, where $N=100$ and $P=160$, using $\alpha = 10/N$ and $\beta = e^{-10}$. The fundamental memory set as well as the initial state $\vetx(0)$ have been generated using an uniform distribution, that is, the entries of the fundamental memories are such that $\Pr[u_i^\xi=1]=\Pr[u_i^\xi=-1]=1/2$ for all $i=1,\ldots,N$ and $\xi=1,\ldots,P$. Figure \ref{fig:EnergyBipolar} shows the evolution of the energy of ECNNs using both synchronous and asynchronous update mode. For a fair comparison, we used $\Delta=1$ and $\Delta t = 1/N$ for synchronous and asynchronous update mode in \eqref{eq:HRCNN}, respectively. Note that the energy,  given by \eqref{eq:energy} with \bb \label{eq:F_e} F(x) = \frac{\beta}{\alpha}e^{\alpha x},\ee
always decreases using both synchronous and asynchronous update. Also, in accordance with Figure \ref{fig:EnergyBipolar}, the bipolar ECNN with asynchronous update mode usually reached an equilibrium earlier than the ECNN with synchronous update. Nevertheless, by taking advantage of matrix operations and parallel computing, our computational implementation of the synchronous ECNN is usually faster than the asynchronous model.
\end{experiment}

\subsection{Complex-Valued Multistate RCNNs}

As far as we know, the first hypercomplex-valued neural networks which have been used to implement associative memories dates to the late 1980s \cite{noest88a,noest88b}. Using the complex-valued multistate signum function \cite{aizenberg92}, Jankowski et al. \cite{jankowski96} proposed a complex-valued Hopfield network which corroborated to the development of many other hypercomplex-valued recurrent neural networks including \citep{lee06,muezzinoglu03,tanaka09,kobayashi17e,kobayashi17f,castro18cnmac,isokawa18}. The complex-valued signum function, as described by \cite{castro20nn,kobayashi17e,kobayashi17f}, is defined as follows: Given a positive integer $K>1$, referred to as the \textit{resolution factor}, define $\Delta \theta = \pi/K$ and the sets
\bb \label{eq:Dmultistate} \mathcal{D}=\{z \in \mathbb{C}\setminus \{0\}:\arg(z) \neq (2k-1) \Delta \theta, \forall k=1,\ldots,K \},\ee  and  
\bb \label{eq:Smultistate} \mathcal{S}=\{1,e^{2\ii \Delta \theta}, e^{4 \ii \Delta \theta},\ldots, e^{ 2(K-1) \ii  \Delta \theta}\}.\ee
Then, the complex-signum function $\csgn:\mathcal{D} \to \mathcal{S}$ is given by the following expression for all $z \in \mathcal{D}$:
\bb \text{csgn}(z) = \begin{cases}
1, & 0 \leq \arg(z) < \Delta \theta, \\
e^{2\ii \Delta \theta}, & \Delta \theta < \arg(z) < 3 \Delta \theta, \\
\quad \vdots & \qquad \qquad \vdots \\
1, & (2K-1) \Delta \theta < \arg(z) < 2 \pi.
             \end{cases} \label{eq:csgn}  
\ee

Using the polar representation of complex numbers, Castro and Valle showed that the complex-valued signum function given by \eqref{eq:csgn} is a $\mathcal{B}$-function in the system of complex numbers with the natural conjugation \cite{castro20nn}. Therefore, from Theorem \ref{thm:Convergence}, we conclude that the complex-valued multistate RCNNs given by \eqref{eq:HRCNN}, \eqref{eq:activation}, and \eqref{eq:weights} with $\phi \equiv \csgn$ always yield a convergent sequence of complex-valued vectors. The following example illustrates the dynamic of a complex-valued multistate ECNN.

\begin{exmp} \label{ex:CvMultistate}
\begin{figure}[t] \label{fig:CvMultistate}
    \centering
    \begin{tabular}{c}
    a) Synchronous update mode \\    \includegraphics[scale=0.8]{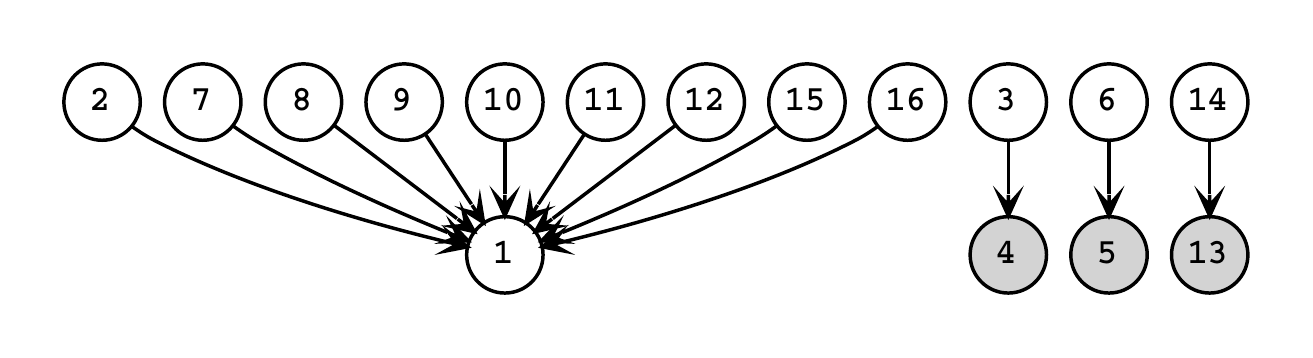} \\
    b) Asynchronous update mode \\
    \includegraphics[scale=0.8]{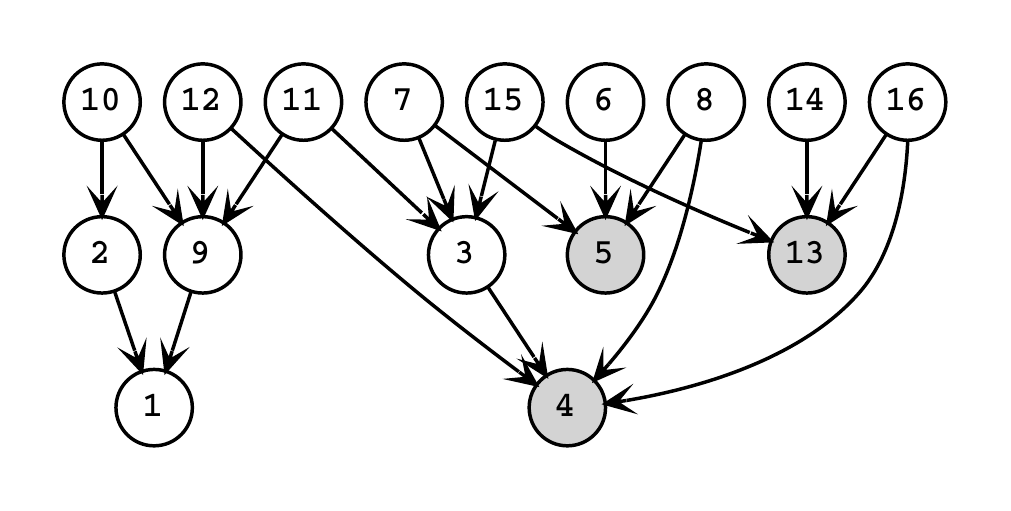}
    \end{tabular}
    \caption{Dynamic of the complex-valued multistate ECNN with $\alpha=\beta=1$ using a) synchronous and b) asynchronous update modes.}
\end{figure}
Consider a resolution factor $K=4$ and the fundamental memory set
\bb \label{eq:UCvMultistate} \mathcal{U} = \left\{
\vetu^1 = \begin{bmatrix} 1 \\ -\ii \end{bmatrix}, 
\vetu^2 = \begin{bmatrix} \ii \\ 1 \end{bmatrix}, 
\vetu^3 = \begin{bmatrix} -\ii \\ 1 \end{bmatrix}
\right\} \subset \mathcal{S}^2,\ee
where $\mathcal{S} = \{1,\ii,-1,-\ii\}$ is the set given by \eqref{eq:Smultistate}. In analogy to Figure \ref{fig:bipolarRCNN}, Figure \ref{fig:CvMultistate} portrays the entire dynamic of the  complex-valued multistate ECNN designed for the storage of the fundamental memory $\mathcal{U}$ given by \eqref{eq:UCvMultistate}. Again, there are 16 possible states and the fundamental memories, which corresponds to the 4th, 5th, and 13th states (the same as in Example \ref{ex:Bipolar}), correspond to gray nodes in Figure \ref{fig:CvMultistate}. In this example, we considered $\alpha=1/2$, $\beta=1$, and both synchronous and asynchronous updates. 
%
%
In agreement with Theorem \ref{thm:Convergence}, the exponential complex-valued multistate RCNN always settled at an equilibrium using either synchronous or asynchronous update. In contrast to the exponential bipolar RCNN presented in Example \ref{ex:Bipolar}, both synchronous and asynchronous multistate models have one spurious memory; the vector $\boldsymbol{s}^1 = [1,1] \in \mathcal{S}^2$, which corresponds to the 1st state. 
\end{exmp}

\begin{experiment}
\begin{figure}[t] \label{fig:EnergyCvMultistate}
    \centering
    \includegraphics[width=0.9\columnwidth]{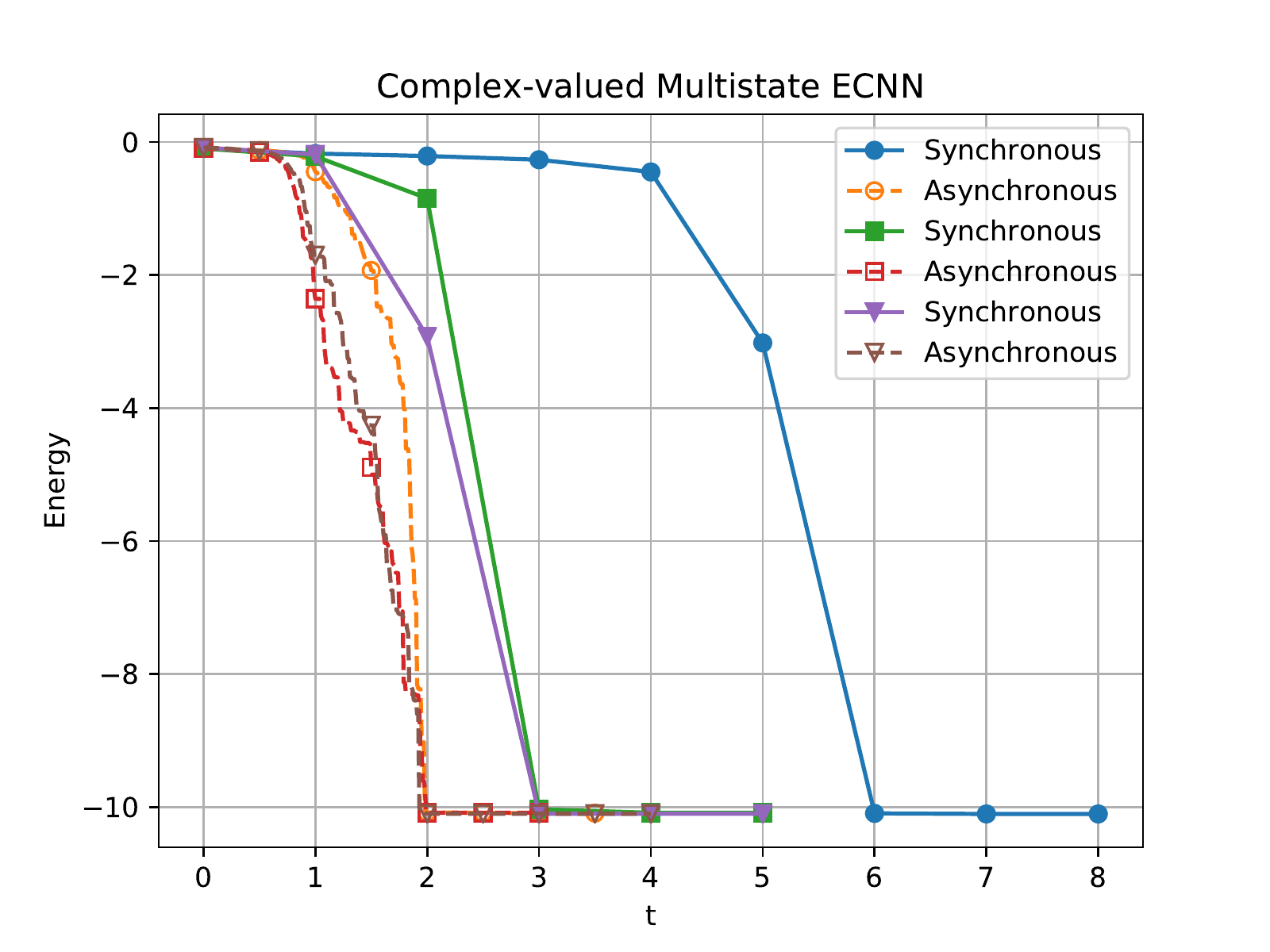}
    \caption{Evolution of the energy of the complex-valued multistate ECNN designed for the storage of $P=160$ uniformly generated fundamental memories of length $N=100$ with a resolution factor $K=256$.}
\end{figure}
Like Experiment \ref{exp:EnergyBipolar}, we synthesized complex-valued multistate ECNN models for the storage and recall of fundamental memories $\mathcal{U} = \{\vetu^1,\ldots,\vetu^P\}$, where $N=100$ and $P = 160$, using $\alpha=10/N$ and $\beta = e^{-10}$. Furthermore, we adopted a resolution factor $K=256$ and randomly generated the components of a fundamental memory, as well as the input patterns, using uniform distribution (the elements of $\mathcal{S}$ given by \eqref{eq:Smultistate} have all equal probability). Figure \ref{fig:EnergyCvMultistate} shows the evolution of the energy of the complex-valued ECNNs operating synchronously and asynchronously. In agreement with Theorem \ref{thm:Convergence}, the energy always decreases until the network settles at an equilibrium. As in the bipolar case, although the ECNNs with asynchronous update usually comes to rest at a stationary state earlier than the ECNN with synchronous update, the computational implementation of the latter is usually faster than the asynchronous model. Furthermore, since the set $\mathcal{D}$ given by \eqref{eq:Dmultistate} is dense on the complex-numbers $\mathbb{C}$, we do not need to check if the argument of $\csgn$ given by \eqref{eq:csgn} belongs to $\mathcal{D}$ in the computational implementation of the complex-valued multistate ECNNs.
\end{experiment}

Finaaly, we would like to point out that complex-valued multistate RCNN are new contributions to the field because the complex-valued recurrent correlation neural networks introduced in \cite{valle14nnB} are based on the continuous-valued function given by $f(z) = z/|z|$ for all complex number $z \neq 0$. 

\subsection{Hyperbolic-Valued Multistate RCNNs}

Neural networks based on hyperbolic numbers constitute an active topic of research since the earlier 2000s  \cite{buchholz00,ontrup01,nitta18}. Hyperbolic-valued Hopfield neural networks, in particular, have been extensively investigated in the last decade by Kobayashi, Kuroe, and collaborators \cite{kuroe11b,kobayashi13,kobayashi16c,kobayashi16d,kobayashi18c,kobayashi19a,kobayashi19c}. Hyperbolic-valued Hopfield neural networks are usually  synthesized using either Hebbian learning \cite{kobayashi13,kobayashi19c} or the projection rule \cite{kobayashi19a}.
On the one hand, like the complex-valued model, the hyperbolic-valued Hopfield neural network with the Hebbian learning suffers from a low storage capacity due to the cross-talk between the stored items \cite{jankowski96,muezzinoglu03,kobayashi19c}. On the other hand, the projection rule may fails to satisfy the stability conditions imposed on the synaptic weights \cite{kobayashi19a}. Examples of the activation function employed on hyperbolic-valued Hopfield neural networks include the split-sign function \cite{kobayashi16c} and the directional multistate activation function \cite{kobayashi18c}. In the following, we address the stability of hyperbolic-valued RCNNs with the directional multistate activation function, which corresponds to the $\csgn$ function given by \eqref{eq:csgn}. 

Such as the complex-numbers, hyperbolic numbers are 2-dimensional hypercomplex numbers of the form $p = p_0 + p_1\ii$. In contrast to the complex imaginary unit, the hyperbolic unit satisfies $\ii^2 = 1$. The set of all hyperbolic numbers is denoted by $\mathbb{U}$. The (natural) conjugate of a hyperbolic number $p = p_0 + p_1\ii$ is $\bar{p} = p_0-p_1\ii$. Using the natural conjugate, the symmetric bilinear form given by \eqref{eq:inner-product} for hyperbolic numbers satisfies $\mathcal{B}(p,q) = p_0q_0 - p_1q_1$, for all $p,q \in \mathbb{U}$. Unfortunately, the $\csgn$ function given by \eqref{eq:csgn} is not a $\mathcal{B}$-function. 
Thus, an hyperbolic-valued RCNN may fail to settle at an equilibrium. The following example and experiment confirm this remark.

\begin{exmp} \label{ex:HvMultistate}
\begin{figure}[t] \label{fig:HvMultistate}
    \centering
    \begin{tabular}{c}
    a) Synchronous update mode \\    \includegraphics[scale=0.8]{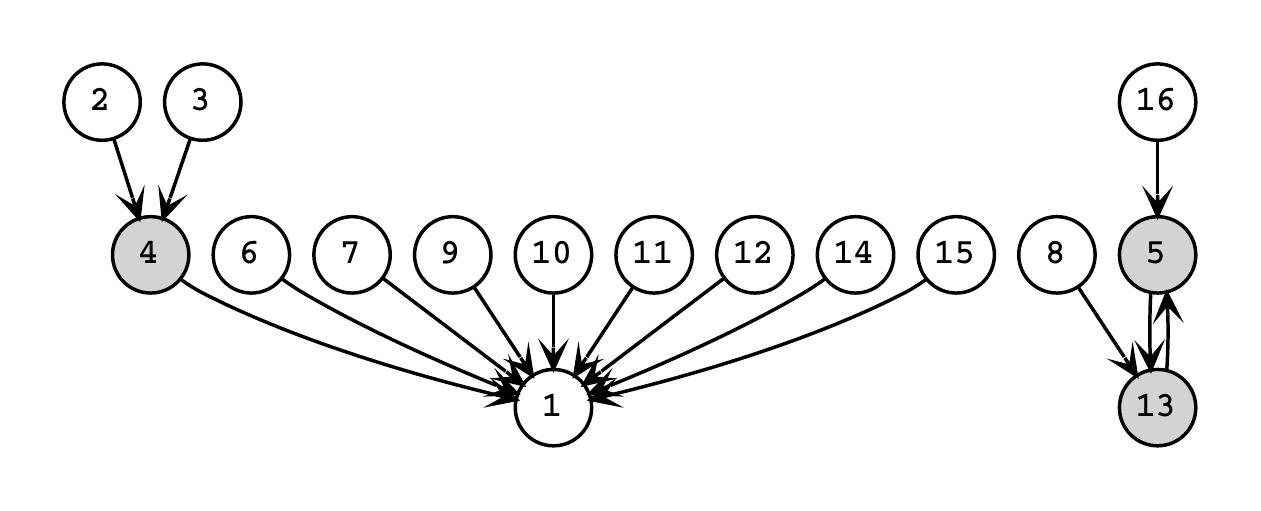} \\
    b) Asynchronous update mode \\
    \includegraphics[scale=0.8]{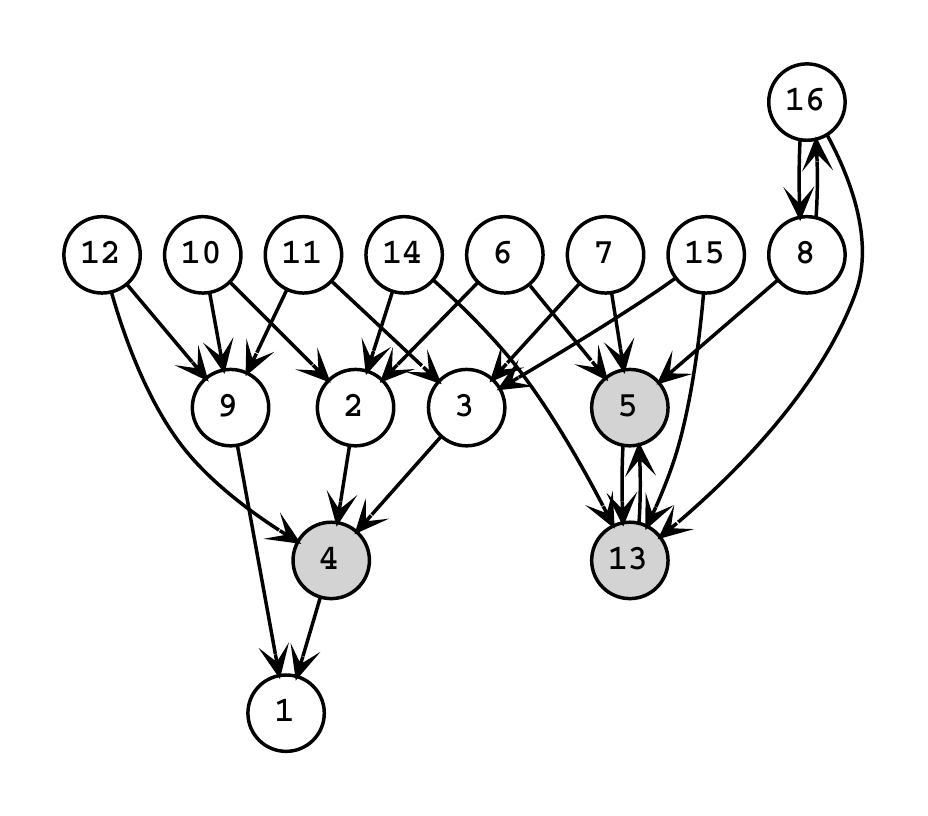}
    \end{tabular}
    \caption{Dynamic of the hyperbolic-valued multistate ECNN with $\alpha=\beta=1$ using a) synchronous and b) asynchronous update modes.}
\end{figure}
Consider the fundamental memory set $\mathcal{U}$ given by \eqref{eq:UCvMultistate}, where $\mathcal{S} = \{1,\ii,-1,-\ii\}$ is the set given by \eqref{eq:Smultistate} and $\ii$ denotes the hyperbolic unit, i.e., $\ii^2=1$. Using $\alpha=1/2$ and $\beta=1$, we synthesized hyperbolic-valued mustistate ECNN models designed for the storage of the fundamental memory set $\mathcal{U}$. Such as in the previous examples, Figure \ref{fig:HvMultistate} depicts the entire dynamic of the hyperbolic-valued ECNN models obtained using synchronous and asynchronous update mode. Note that the network, using either synchronous or asynchronous update, may cycle between the 5th and 13th states, which correspond respectively to the fundamental memories $\vetu^2$ and $\vetu^3$. 
\end{exmp}

\begin{experiment}
\begin{figure}[t] \label{fig:EnergyHvMultistate}
    \centering
    \includegraphics[width=0.9\columnwidth]{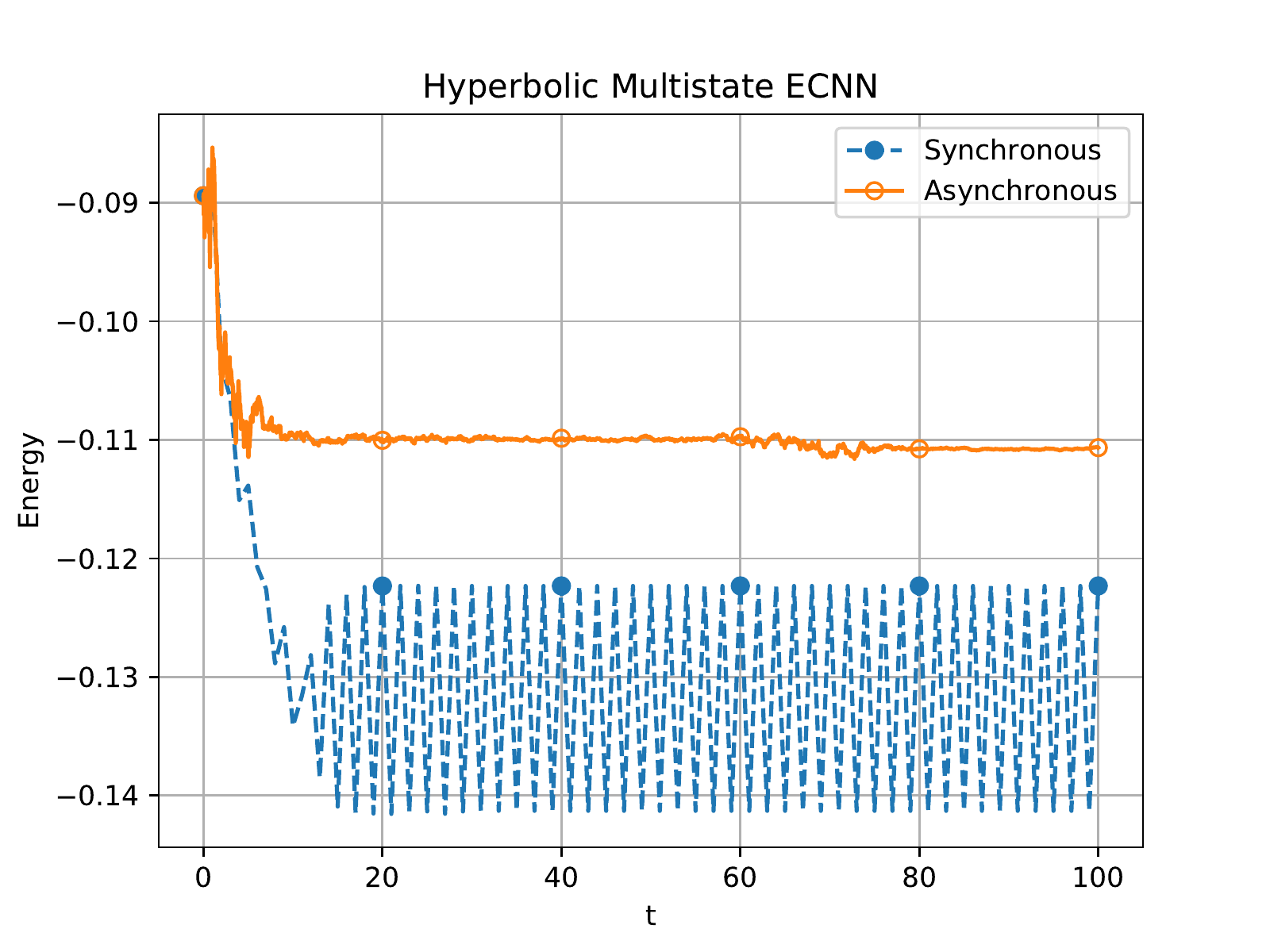} 
    \caption{Evolution of the energy of the hyperbolic-valued multistate ECNN designed for the storage of $P=160$ uniformly generated fundamental memories of length $N=100$ with a resolution factor $K=256$.}
\end{figure}
Like in the previous experiments, we synthesized hyperbolic-valued ECNN models using $\alpha=10/N$ and $\beta=e^{-10}$ designed for the storage of uniformly distributed fundamental memories $\vetu^1,\ldots,\vetu^{160}$ of length $N=100$ with components in the multivalued set $\mathcal{S}$ given by \eqref{eq:Smultistate} with $K=256$. The multistate ECNNs have been initialized with a random state $\vetx(0)$ uniformly distributed in $\mathcal{S}^N$. Figure \ref{fig:EnergyHvMultistate} shows the evolution of the energy of an hyperbolic-valued multistate model using synchronous and asynchronous update modes. In both cases, the hypercomplex-valued ECNN failed to settle at an equilibrium. 
\end{experiment}

We would like to point out that we can ensure the stability of hyperbolic-valued multistate RCNNs by either considering a different reverse-involution or a different activation function. In fact, by considering the trivial reverse involution $\tau(p)=p$ instead the natural conjugation $\tau(p) = \bar{p}$, the hyperbolic-valued RCNNs coincide with the complex-valued RCNN with the natural conjugation. Alternatively, \eqref{eq:HRCNN} yields a convergent sequence of multivalued vectors if we adopt the function $\overline{\csgn}:\mathcal{D} \to \mathcal{S}$ defined by $\overline{\csgn}(p) = \csgn(\bar{p})$ for all $p \in \mathcal{D}$. In this case, however, the fundamental memories may fail to be stationary states.

\begin{exmp} \label{ex:cHvMultistate}
\begin{figure}[t] \label{fig:cHvMultistate}
    \centering
    \begin{tabular}{c}
    a) Synchronous update mode \\    \includegraphics[scale=0.8]{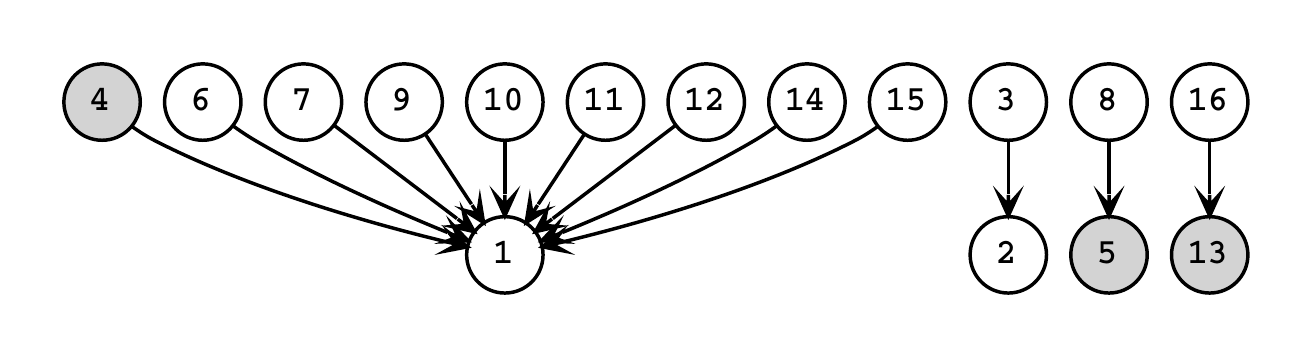} \\
    b) Asynchronous update mode \\
    \includegraphics[scale=0.8]{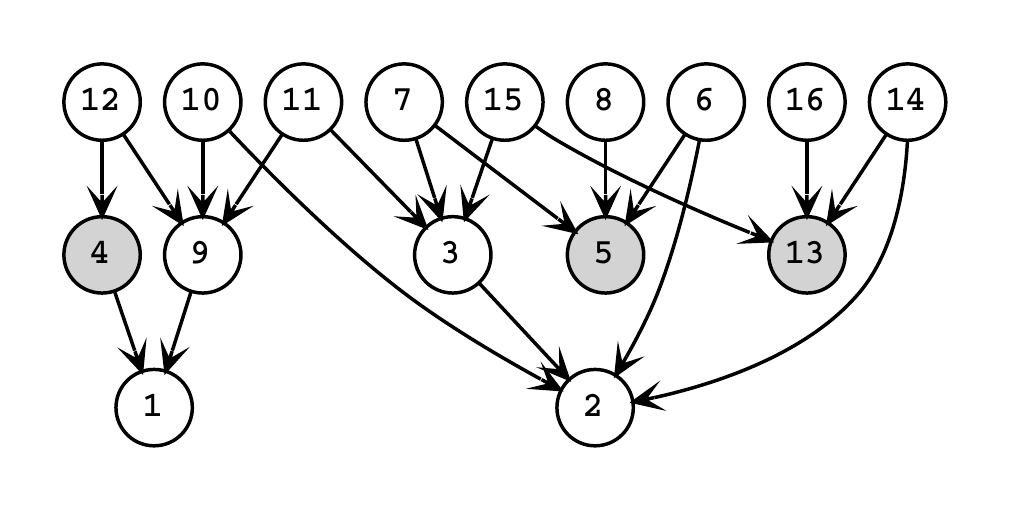}
    \end{tabular}
    \caption{Dynamic of the hyperbolic-valued ECNN with $\alpha=\beta=1$ and the activation function $\overline{\csgn}(p) = \csgn(\bar{p})$ using a) synchronous and b) asynchronous update modes.}
\end{figure}
We synthesized an hyperbolic-valued ECNN designed for the storage of the fundamental memory set $\mathcal{U}$ given by \eqref{eq:UCvMultistate} using $\alpha=\beta=1$ and the activation function $\overline{\csgn}$. Figure \ref{fig:cHvMultistate} depicts the entire dynamic of this hyperbolic-valued ECNN using synchronous and asynchronous update mode. In contrast to the model in Example \ref{ex:HvMultistate}, this hyperbolic-valued ECNN always settle at an equilibrium. However, the fundamental memory $\vetu^1$, which corresponds to the 4th state, is not a stationary state.
\end{exmp}

Finally, we would like to recall that both complex and hyperbolic number systems are isomorphic to instances of real Clifford algebras. Apart from complex and hyperbolic-valued models, Hopfield neural networks on Clifford algebras have been investigated by Vallejo and Bayro-Corrochano \cite{vallejo08}, Kuroe and collaborators \cite{kuroe11b,kuroe13}, and Kobayashi \cite{kobayashi18b}. A brief account on the stability of Hopfield neural networks on Clifford algebras of dimension 2 with the so-called split-sign activation function can be found in \cite{castro20nn}. Following the reasoning presented in this section and in \cite{castro20nn}, we believe one can easily investigate the stability of complex-valued, hyperbolic-valued, and dual number-valued RCNNs with the split-sign activation function. 

\subsection{Quaternion-Valued Multistate RCNNs}

Quaternions constitute a four-dimensional associative but non-commutative hypercomplex number system. Formally, a quaternion is a hypercomplex number of the form $p = p_0 + p_1\ii+p_2\jj + p_3 \kk$ in which the hypercomplex units\footnote{As usual, we denote the quaternionic hypercomplex units by $\ii$, $\jj$, and $\kk$ instead of $\ii_1$, $\ii_2$, and $\ii_3$, respectively.} satisfy $\ii^2 = \jj^2 = \kk^2 = -1$ and $\ii \jj = \kk$. Apart from representing three and four-dimensional data as a single entity, quaternions are particularly useful to describe rotations in the three dimensional space \cite{arena98,buelow99,kuipers99}. As a growing and active research area, quaternion-valued neural networks have been successfully applied for signal processing and times series prediction \cite{arena97,xia15,xu16,papa17,xiaodong17,greenblatt18} as well as image processing and classification \cite{minemoto16,minemoto17,castro17bracis,chen17,shang14}. Furthermore, quaternion-valued outperformed their corresponding real-valued models in many of the aforementioned applications. 

In the context of recurrent networks and associative memories, research on quaternion-valued Hopfield neural network dates to late 2000s \cite{isokawa07,isokawa08a,isokawa08b}. Precisely, based on the works of Jankowski et al. \cite{jankowski96}, Isokawa et al. introduced a quaternion-valued multisate signum function using quantizations of the phase-angle representation of quaternions \cite{isokawa08b,buelow99}. The quaternion-valued multisate signum function have been used to define multistate Hopfield network which can be synthesized using the Hebbian learning or the projection rule \cite{isokawa08b,isokawa13}. Unfortunately, the quaternion-valued multistate signum function is not a $\mathcal{B}$-function on the quaternions with the natural conjugation \cite{castro20nn}. Therefore, we cannot ensure the stability of a quaternion-valued Hopfield network based on the quaternion-valued multistate signum function. In fact, there are simple examples in which this network does not come to rest at an equilibrium \cite{valle18tnnls}. The continuous-valued and the twin-multistate activation functions provide alternatives to define stable quaternion-valued Hopfield neural networks \cite{valle14bracis,valle18tnnls,kobayashi16c,kobayashi17c}.

The continuous-valued activation function $\sigma$ is defined by  \bb \label{eq:sigmaQ} \sigma(q) = \frac{q}{|q|}, \ee for all non-zero quaternion $q = q_0+q_1\ii+q_2\jj+q_3\kk$, where $|q| = \sqrt{q_0^2+q_1^2+q_2^2+q_3^2}$ denotes the absolute value or norm of $q$ \cite{aizenberg92,valle14bracis}. From the Cauchy-Schwartz inequality, one can show that $\sigma$ is a $\mathcal{B}$-function in the set of quaternions with the natural conjugation. From Theorem \ref{thm:Convergence}, and in accordance with \cite{valle18wcci}, the sequence produced by a quaternion-valued RCNN with the continuous-valued activation function $\sigma$ is always convergent. The storage capacity and noise tolerance of the continuous-valued quaternionic ECNN have been investigated on \cite{valle18wcci}. Let us now turn our attention to the new quaternion-valued multistate RCNN obtained by considering the twin-multistate activation function.

The twin-multistate activation function, introduced by Kobayashi \cite{kobayashi17a}, is defined using the complex-valued multistate signum function given by \eqref{eq:csgn} \cite{kobayashi17a}. Precisely, from the identity $\ii \jj = \kk$, we can write a quaternion as 
\bb \label{eq:quat2complex} q = (q_0+q_1\ii)+(q_2+q_3\ii)\jj =  z_0 + z_1 \jj, \ee
where $z_0 = q_0+q_1\ii$ and $z_1 = q_2+q_3\ii$ are complex numbers. Given a resolution factor $K$, the twin-multistate activation function is defined by means of the equation 
\bb \label{eq:tcsgn} 
\tsgn(q) = \csgn(z_0) + \csgn(z_1)\jj, \ee
for all $q = z_0+z_1\jj$ such that $z_0,z_1 \in \mathcal{D}$, where $\mathcal{D}$ is the domain of the complex-valued multistate signum function defined by \eqref{eq:Dmultistate}. Note that the domain $\mathcal{D}_t$ and codomain $\mathcal{S}_t$ of the twin-multistate activation function $\tsgn$ are respectively the subset of quaternions 
\bb \label{eq:D_t} \mathcal{D}_t = \{q = z_0+z_1\jj: z_0,z_1 \in \mathcal{D}\}, \ee and 
\bb \label{eq:S_t} \mathcal{S}_t = \{s = w_0+w_1\jj: w_0,w_1 \in \mathcal{S}\},\ee
where $\mathcal{S}$ is the set complex numbers defined by \eqref{eq:Smultistate}. It is not hard to show that $\mathcal{S}_t$ is compact. Furthermore, $\tsgn$ is a $\mathcal{B}$-function on the quaternions with the natural conjugation. In fact, using \eqref{eq:quat2complex}, we can express the quaternion-valued symmetric bilinear form $\mathcal{B}_\mathbb{Q}$ as the sum of the complex-valued symmetric bilinear form $\mathcal{B}_\mathbb{C}$ as follows for any quaternions $q = z_0+z_1\jj$ and $p = w_0+w_1\jj$:
\bb \mathcal{B}_\mathbb{Q}(q,s) =  \mathcal{B}_\mathbb{C}(z_0,w_0)+ \mathcal{B}_\mathbb{C}(z_1,w_1).\ee  
Since $\csgn$ is a $\mathcal{B}$-function on the complex-numbers with the natural conjugation, we obatin the following inequality for all $q  = z_0+z_1\jj \in \mathcal{D}_t$ and $s = w_0+w_1\jj \in \mathcal{S}_t \setminus \{\tsgn(q)\}$:
\begin{align}
    \mathcal{B}_\mathbb{Q}(\tsgn(q),q) &= \mathcal{B}_\mathbb{C}(\csgn(z_0),z_0)+\mathcal{B}_\mathbb{C}(\csgn(z_1),z_1) \\ &>\mathcal{B}_\mathbb{C}(w_0,z_0)+\mathcal{B}_\mathbb{C}(w_1,z_1) \\ &= \mathcal{B}_\mathbb{Q}(s,q).  
\end{align}
From Theorem \ref{thm:Convergence}, we conclude quaternion-valued RCNNs with twin-multistate activation functions always settle at a stationary state. The following example and computational experiment confirm this remark.

\begin{exmp}
Using a resolution factor $K=4$, we obtain the set 
\bb \mathcal{S}_t = \{\pm 1  \pm \jj,\pm 1 \pm \kk, \pm\ii \pm\jj,\pm\ii\pm\kk  \}. \ee
Let us consider the fundamental memory set 
\bb \label{eq:Uquat} \mathcal{U} = \{\vetu^1 = 1-\kk, \quad \vetu^2 = \ii+\jj, \quad \vetu^3 = -\ii+\jj \}. \ee 
We synthesized the quaternion-valued multistate ECNN designed for the storage and recall of the fundamental memory set $\mathcal{U}$ given by \eqref{eq:Uquat}. Like the previous experiment, we used $\alpha=1/2$ and $\beta=1$. In this simple example, the update mode is irrelevant because the network has a single state neuron. Moreover, the dynamic of this quaternion-valued multistate ECNN is identical to the sychronous complex-valued multistate neural network depicted in Figure \ref{fig:CvMultistate}a). 
\end{exmp}

\begin{experiment}
\begin{figure}[t] \label{fig:QvMultistate}
    \centering
    \includegraphics[width=0.9\columnwidth]{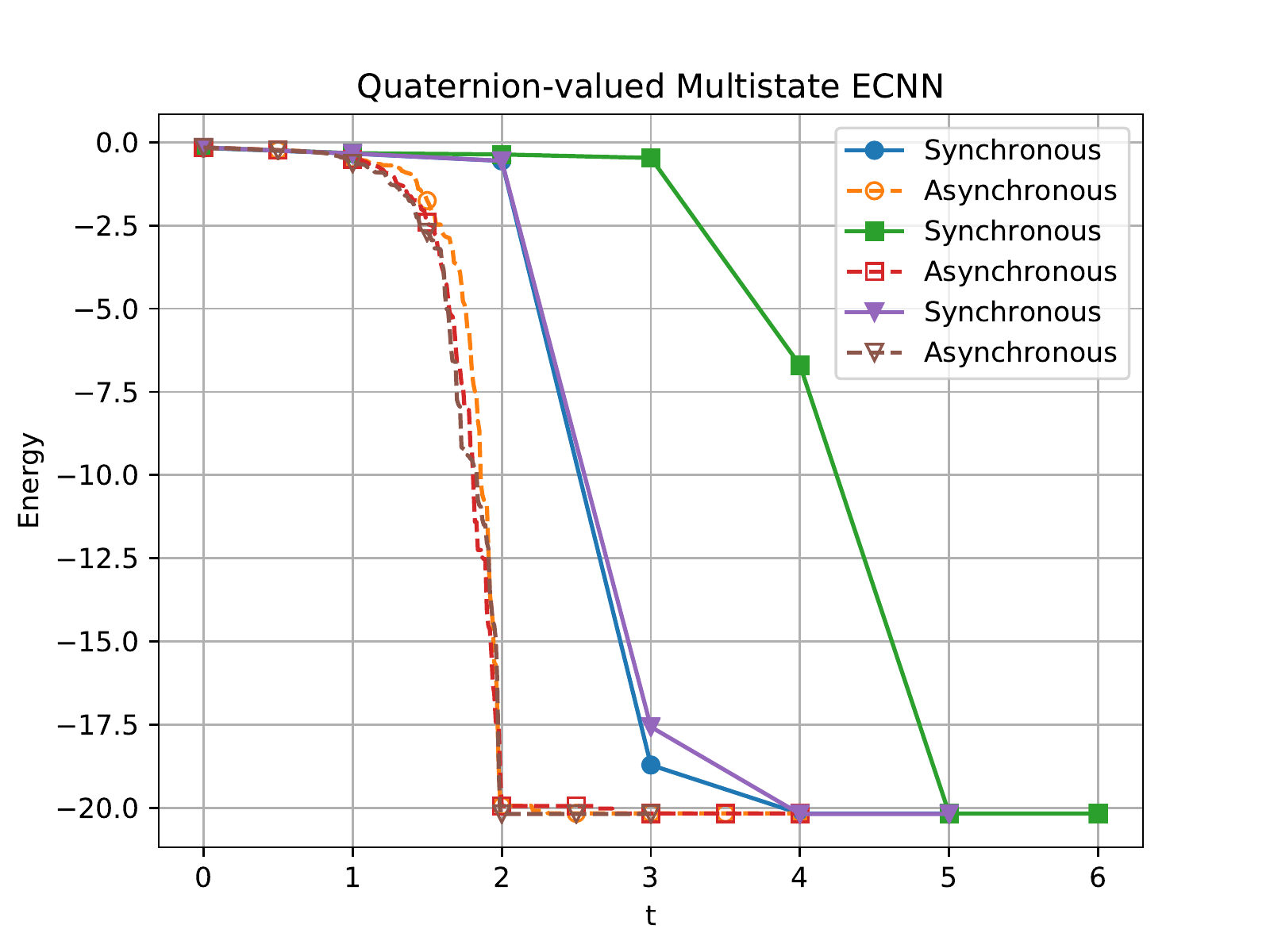} 
    \caption{Evolution of the energy of the quaternion-valued multistate ECNN designed for the storage of $P=160$ uniformly generated fundamental memories of length $N=100$ and $K=16$.}
\end{figure}
We synthesized a quaternion-valued multistate ECNN designed for the storage and recall of randomly generated fundamnetal memory sets $\mathcal{U}=\{\vetu^1,\ldots,\vetu^P\} \subset \mathcal{D}_t^N$, where $N=100$ and $P=160$. In this experiment, we considered $\alpha = 10/(2N)$, $\beta = e^{-10}$, and $K=16$. Furthermore, since the set $\mathcal{D}_t$ is dense on $\mathbb{Q}$, we refrained to check the condition $h_i(t) \in \mathcal{D}_t$ in our computational implementation. Figure \ref{fig:QvMultistate} shows the evolution of the energy of the quaternion-valued multistate ECNN using synchronous and asynchronous updates. Note that the energy always decreases until the networks settle at a stationary state using both update modes. Like the previous hypercomplex-valued ECNN models, the computational implementation of the synchronous quaternion-valued multistate ECNN is usually faster than the asynchronous version. 
\end{experiment}

\subsection{Hypercomplex-valued RCNNs on Cayley-Dickson Algebras}

Complex numbers, quaternions, and octonions are instance of Cayley-Dickson algebras. Cayley-Dickson algebras are hypercomplex number systems of dimension $2^k$ in which the product and the conjugation (reverse involution) are defined recursively as follows: First, let $A_0$ denote the real number system. Given a Cayley-Dickson algebra $A_k$, the next algebra $A_{k+1}$ comprises all pairs $(x,y) \in A_k \times A_k$ with the component-wise addition, the conjugation given by
\bb \label{eq:CayleyConjug} \overline{(x,y)} = (\bar{x},-y),\ee 
and the product defined by 
\bb \label{eq:CayleyProd} (x_1,y_1)(x_2,y_2) = (x_1x_2 - y_2\bar{y}_1, \bar{x}_1 y_2 + x_2 y_1),\ee 
for all $(x_1,y_1),(x_2,y_2) \in A_{k+1}$. The Cayley-Dickson algebras $A_1$, $A_2$, and $A_3$ coincide respectively with the complex numbers, quaternions, and octonions. The symmetric bilinear form $\mathcal{B}:A_k \times A_k \to \mathbb{R}$ is given by 
\bb \label{eq:CaylaeyB} \mathcal{B}(p,q) = \sum_{j=1}^n p_j q_j,\ee 
for all $p = \hyper{p}{n} \in A_k$ and $q = \hyper{q}{n} \in A_k$, where $n=2^k-1$. Note that the symmetric bilinear form $\mathcal{B}$ given by \eqref{eq:CaylaeyB} corresponds to the inner product between $p \equiv (p_0,\ldots,p_n)$ and $q \equiv (q_0,\ldots,q_n)$. As a consequence, Cayley-Dickson algebras enjoy many properties from Euclidean spaces, including the Cauchy-Schwarz inequality. On the downside, the Cayley-Dickson algebras are non-commutative for $k \geq 2$. Thus, particular attention must be given to the order of the terms in \eqref{eq:CayleyProd}. Furthermore, Cayley-Dickson algebras are non-associative for $k \geq 3$. As a consequence, in contrast to complex numbers and quaternions, the product defined by \eqref{eq:CayleyProd} cannot be represented by matrix operations for $k \geq 3$.

In \cite{castro20nn}, de Castro and Valle show that the continuous-valued function $\sigma$ given by \eqref{eq:sigmaQ} is a $\mathcal{B}$-function in any Cayley-Dickson algebra $A_k$, with $|p|=\sum_{j=1}^n p_j^2$ for all $p = \hyper{p}{n}$, $n=2^k-1$. From Theorem \ref{thm:Convergence}, hypercomplex-valued RCNNs with the continuous-valued function $\sigma$ always settle at an equilibrium synchronously as well as asynchronously. This result generalizes the stability analysis of the complex-valued and quaternion-valued RCNNs introduced in \cite{valle14nnB,valle18wcci}.

Apart from the continuous-valued RCNNs, the split sign function is also a $\mathcal{B}$-function in Cayley-Dickson algebras \cite{castro20nn}. The split sign function $\sgn:\mathcal{D}_s \to \mathcal{S}_s$ is defined by means of the following equation for all $p=\hyper{p}{n} \in A_k$:
\bb \label{eq:split-sign} \sgn(p) = \sgn(p_0)+\sgn(p_1) \ii_1 + \ldots + \sgn(p_n)\ii_n,\ee 
where 
\bb \label{eq:D_s} \mathcal{D}_s = \{p \in A_k: p_0 p_1 \ldots p_n \neq 0\},\ee
and
\bb \label{eq:S_s} \mathcal{S}_s = \{p \in A_k: p_j \in \{-1,+1\}, \forall j=0,1,\ldots,n\}.\ee
From Theorem \ref{thm:Convergence}, the sequence produced by hypercomplex-valued RCNNs with $\phi = \sgn$ are all convergent using synchronous and asynchronous update mode. The following experiment confirms this remark.

\begin{experiment}
\begin{figure}[t] \label{fig:EnergyOctonionSplit}
    \centering
    \includegraphics[width=0.9\columnwidth]{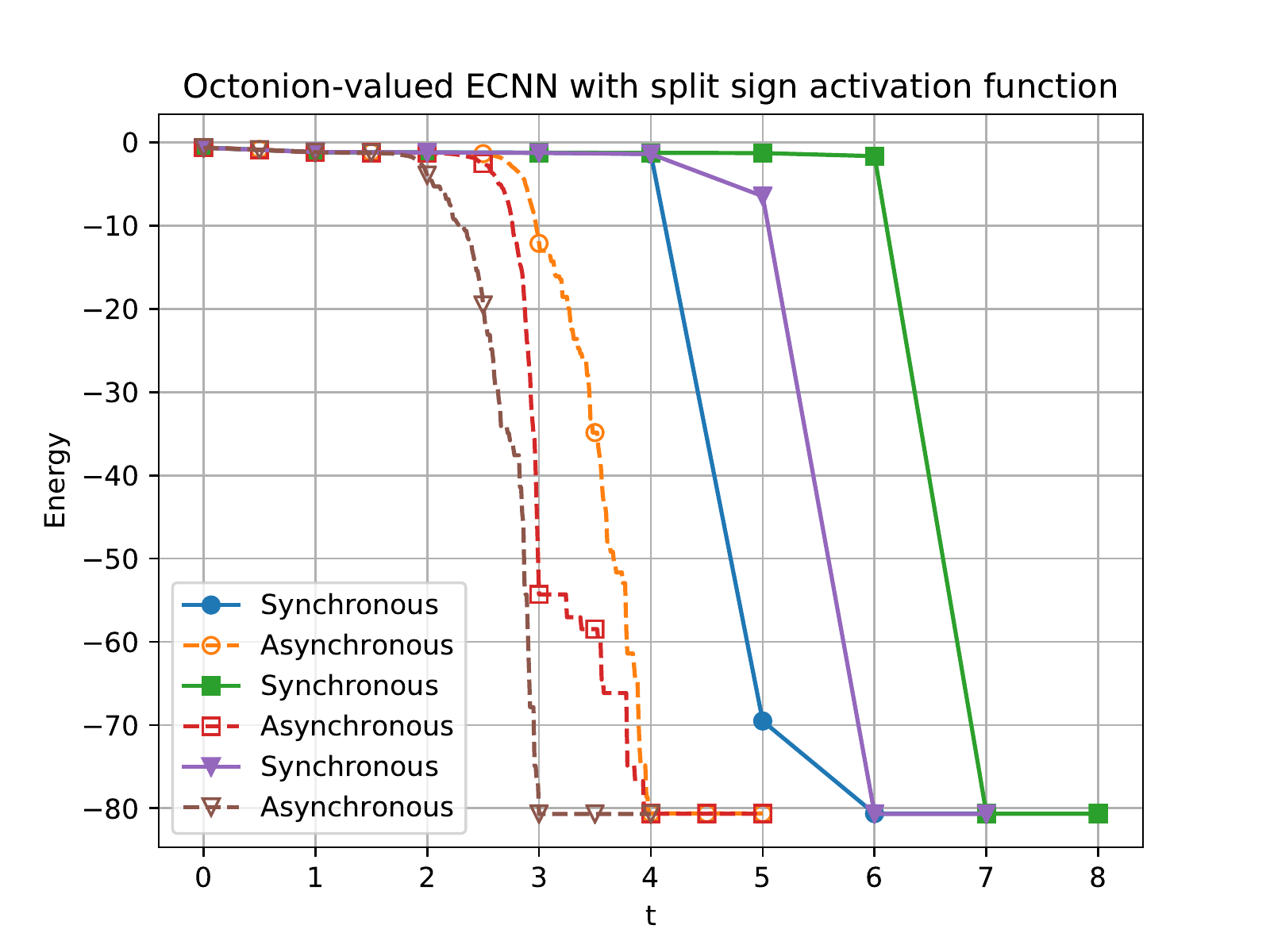} 
    \caption{Evolution of the energy of the octonion-valued ECNN with the split sign function designed for the storage of $P=160$ uniformly generated fundamental memories of length $N=100$.}
\end{figure}
As an illustractive example of a hypercomplex-valued RCNN on Cayley-Dickson algebra, let us consider an octonion-valued ECNN with the split sign activation function. Precisely, we synthesized octonion-valued ECNNs designed form the storage and recall of fundamental memory sets $\mathcal{U} = \{\vetu^1,\ldots,\vetu^P\} \in \mathcal{S}_s^N$, where $\mathcal{S}_s$ is given by \eqref{eq:S_s}, with $N=100$ and $P=160$. In this experiment, we adopted the parameters $\alpha = 10/(8N)$ and $\beta = e^{-10}$. The fundamental memories as well as the initial states have been generated using an uniform distribution, that is, the entries $u_i^\xi = u_{i0}^\xi + u_{i1}^\xi \ii_1 + \ldots + u_{i7}^\xi \ii_7$ of the fundamental memories are such that $\Pr[u_{ij}^\xi=-1]=\Pr[u_{ij}^\xi=+1] = 1/2$ for all $\xi=1,\ldots,P$, $i=1,\ldots,N$, and $j=0,1,\ldots,7$. Figure \ref{fig:EnergyOctonionSplit} shows the evolution of the energy of some octonion-valued ECNN models using both synchronous and asynchronous update. As expected, using both update modes, the energy decreases until the network settles at a stationary state.
\end{experiment}

\subsection{Storage and Recall of Gray-Scale Images}

In the previous subsections, we provided several examples of hypercomplex-valued ECNNs and, by means of simple computational experiments, validated Theorem \ref{thm:Convergence}. Let us now compare the noise tolerance of the hypercomplex-valued ECNNs designed for the storage and recall of gray-scale images.

The CIFAR dataset contain thousands color images of size $32 \times 32$. We randomly selected $P=200$ images from the CIFAR dataset, converted them to gray-scale images, and encoded them in an appropriate manner as hypercomplex-valued vectors. Precisely, we encoded the gray-scale images into hypercomplex-valued vectors as follows where $x \in \{0,1,\ldots,255\}$ denotes a gray-scale pixel value:
\begin{itemize}
    \item \textit{Bipolar:} Using the binary representation $b_1\ldots b_8$, a gray-scale pixel value $x$ can be associated to an array $[2b_1-1,\ldots,2b_8-1] \in \{-1,+1\}^8$. Therefore, a gray-scale image of size $32\times 32$ can be converted into a bipolar vector of length $N_b=8192$ by concatenating all 8-dimensional bipolar arrays. 
    \item \textit{Complex-valued multistate:} Using a resolution factor $K=256$, a gray-scale value $x \in \{0,\ldots,255\}$ yields a complex number $z \in \mathcal{S}$, where $\mathcal{S}$ is the set given by \eqref{eq:Smultistate}, by means of the equation $z = e^{2\pi x \ii/K}$. As a consequence, a gray-scale image of size $32 \times 32$ corresponds to a complex-valued vector in $\mathcal{S}^{N_c}$, where $N_c=1024$.
    \item \textit{Quaternion-valued multistate:} The binary representation $b_1\ldots b_8$ of $x \in \{0,1,\ldots,255\}$ allow us to define integers $x_1 = 8b_4+4b_3+2b_2+b_1$ and $x_2 = 8b_8+4b_7+2b_6+b_5$. Using a resolution factor $K=16$, the equation $q = e^{2\pi x_1 \ii/K} + e^{2\pi x_2 \ii/K} \jj$ yields a quaternion in the set $\mathcal{S}_t$ given by \eqref{eq:S_t}. In this way, a gray-scale image of size $32 \times 32$ corresponds to quaternion-valued vector in $\mathcal{S}_t^{N_q}$, where $N_q=1024$.
    \item \textit{Octonion-valued:} Using the binary representation $b_1\ldots b_8$ of an integer $x \in \{0,\ldots,255\}$, the equation $p = (2b_1-1)+(2b_2-1)\ii_1 + \ldots + (2b_8-1)\ii_7$ yields an octonion $p \in \mathcal{S}_s$, where $\mathcal{S}_s$ is given by \eqref{eq:S_s}. As a consequence, a gray-scale image of size $32 \times 32$ corresponds to an octonion-valued vector in $\mathcal{S}_s^{N_o}$, where $N_o = 1024$.
\end{itemize}
\begin{table}[t]
    \centering
    \begin{tabular}{||c|c|c|c|c||} \hline \hline
         & Bipolar & Complex & Quaternion & Octonion \\ \hline \hline
    $\alpha$ & $20/(N_b)$ & $20/N_c$ & $20/(2N_q)$ & $20/(8N_o)$ \\ 
     $\beta$ & $e^{-20}$ & $e^{-20}$ & $e^{-20}$ & $e^{-20}$ \\ $K$ & --- & 256 & 16 & ---
     \\ \hline \hline
    \end{tabular}
    \caption{Parameters of the hypercomplex-valued ECNN models designed for the storage and recall of CIFAR images where $N_b = 8192$ and $N_c=N_q=N_o = 1024$.}
    \label{tab:parameters}
\end{table}
\begin{figure}[t]
    \centering
    \begin{tabular}{ccc}
    a) Original image & 
    b) Corrupted image &
    c) Bipolar \\
            \includegraphics[width=0.3\columnwidth]{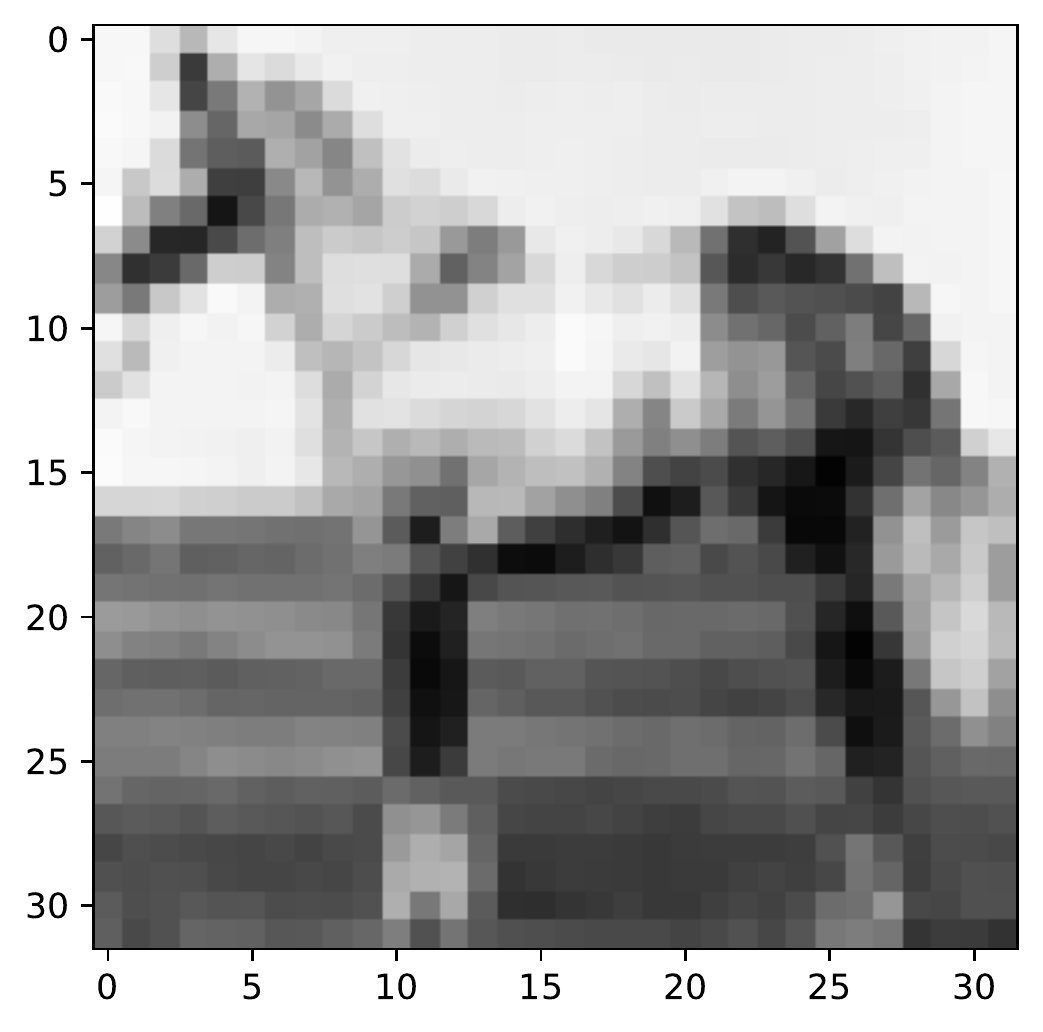} & 
            \includegraphics[width=0.3\columnwidth]{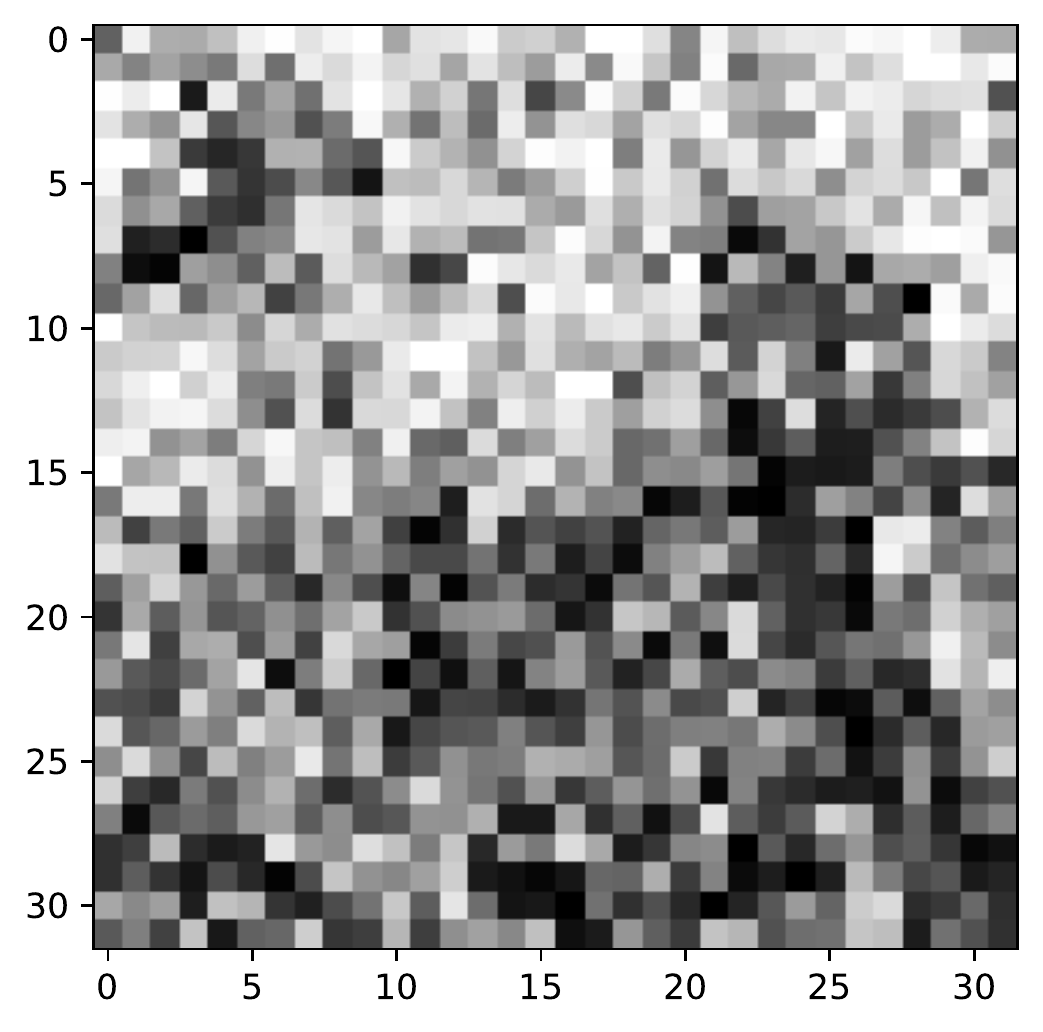} &
            \includegraphics[width=0.3\columnwidth]{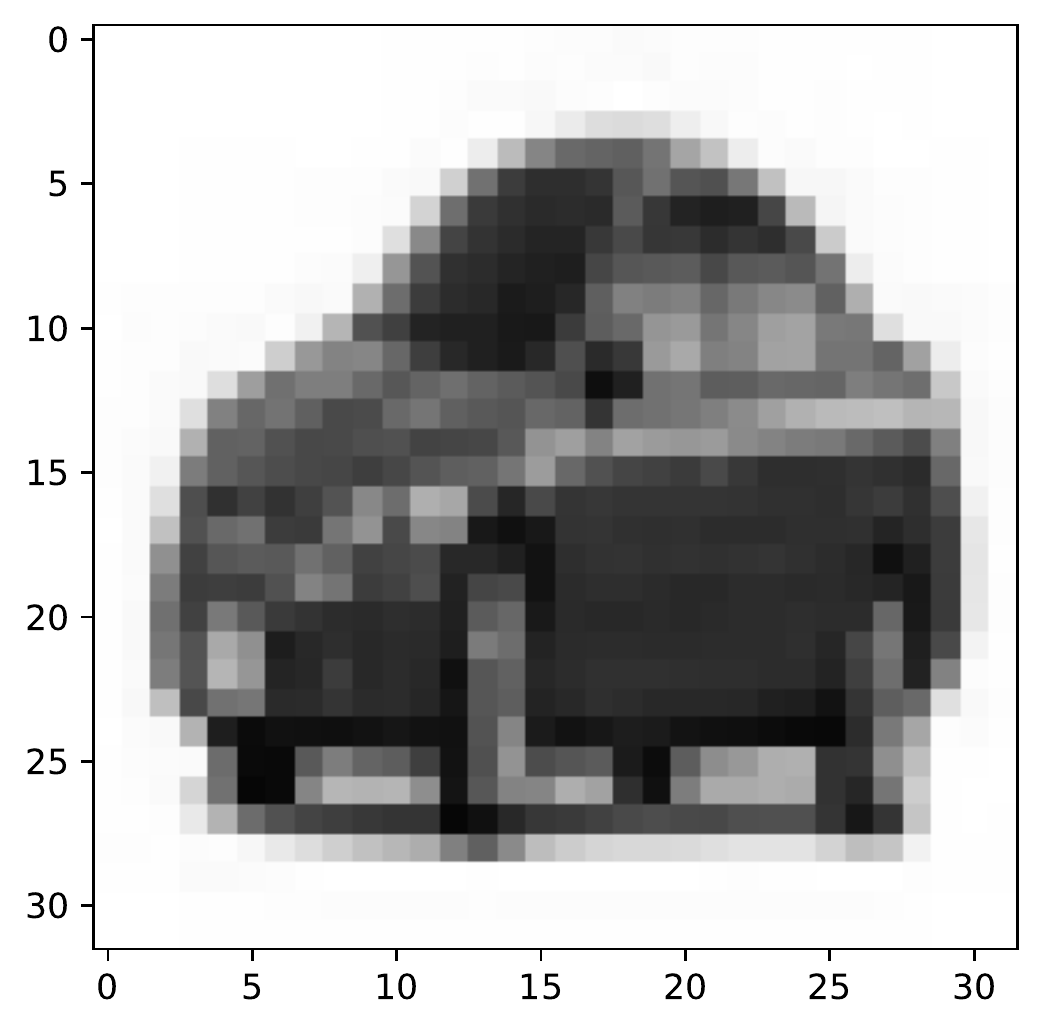} \\
    d) Complex-valued & 
    e) Quaternion-valued &
    f) Octonion-valued \\
            \includegraphics[width=0.3\columnwidth]{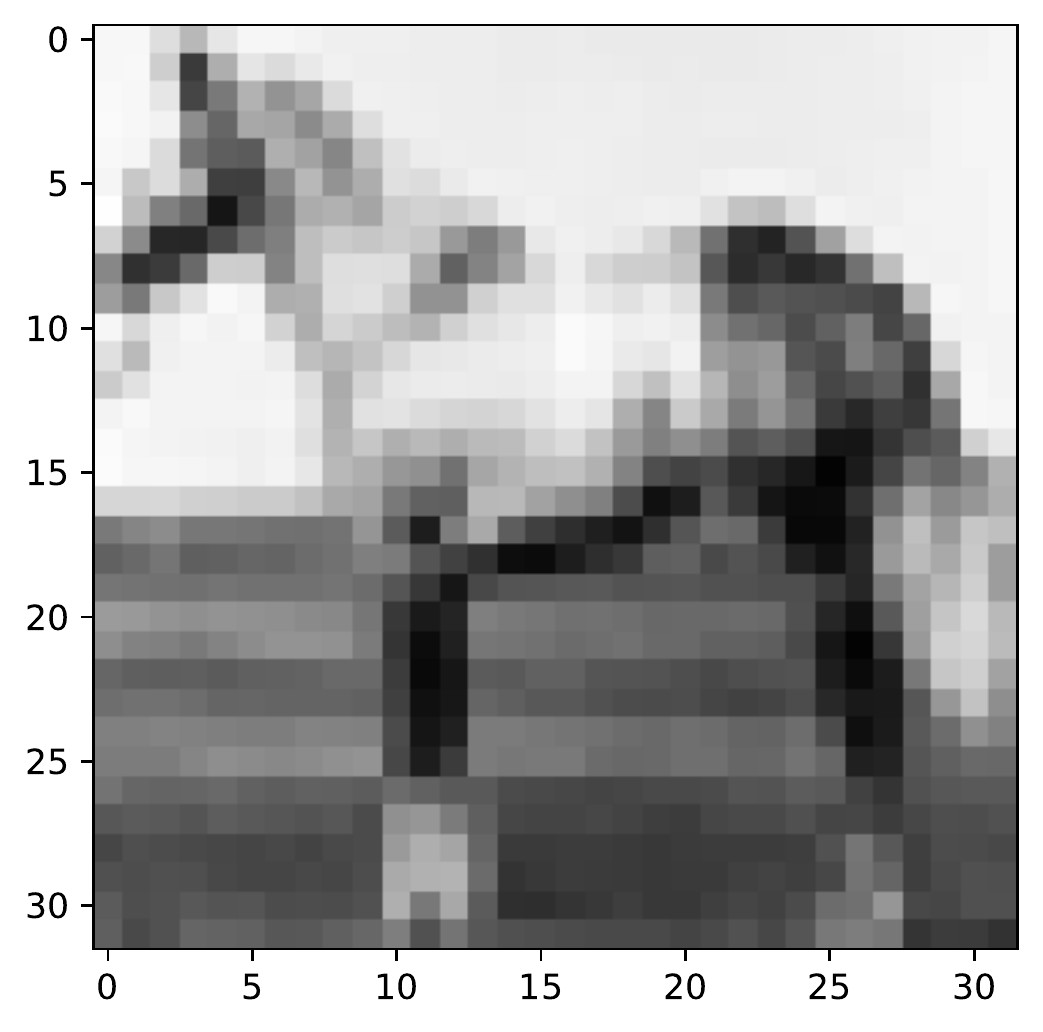} & 
            \includegraphics[width=0.3\columnwidth]{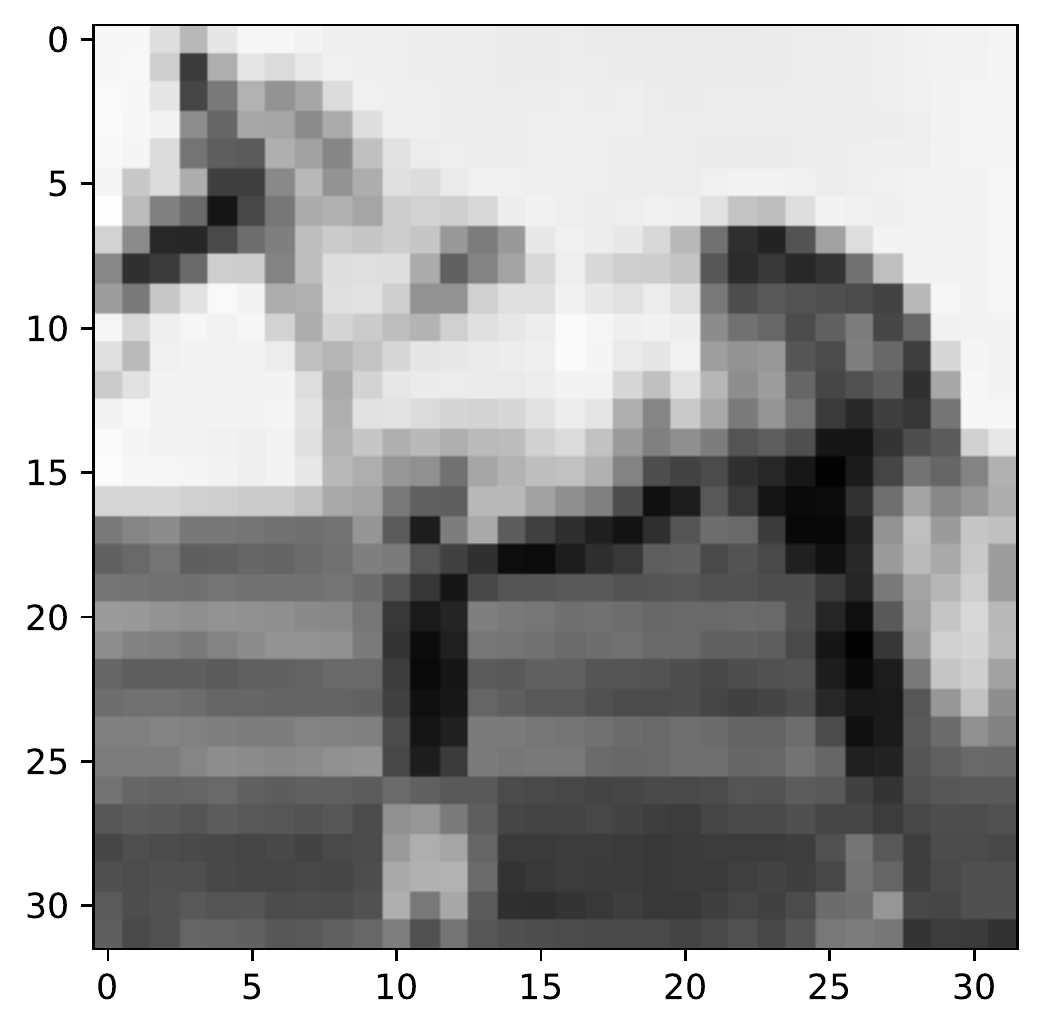} &
            \includegraphics[width=0.3\columnwidth]{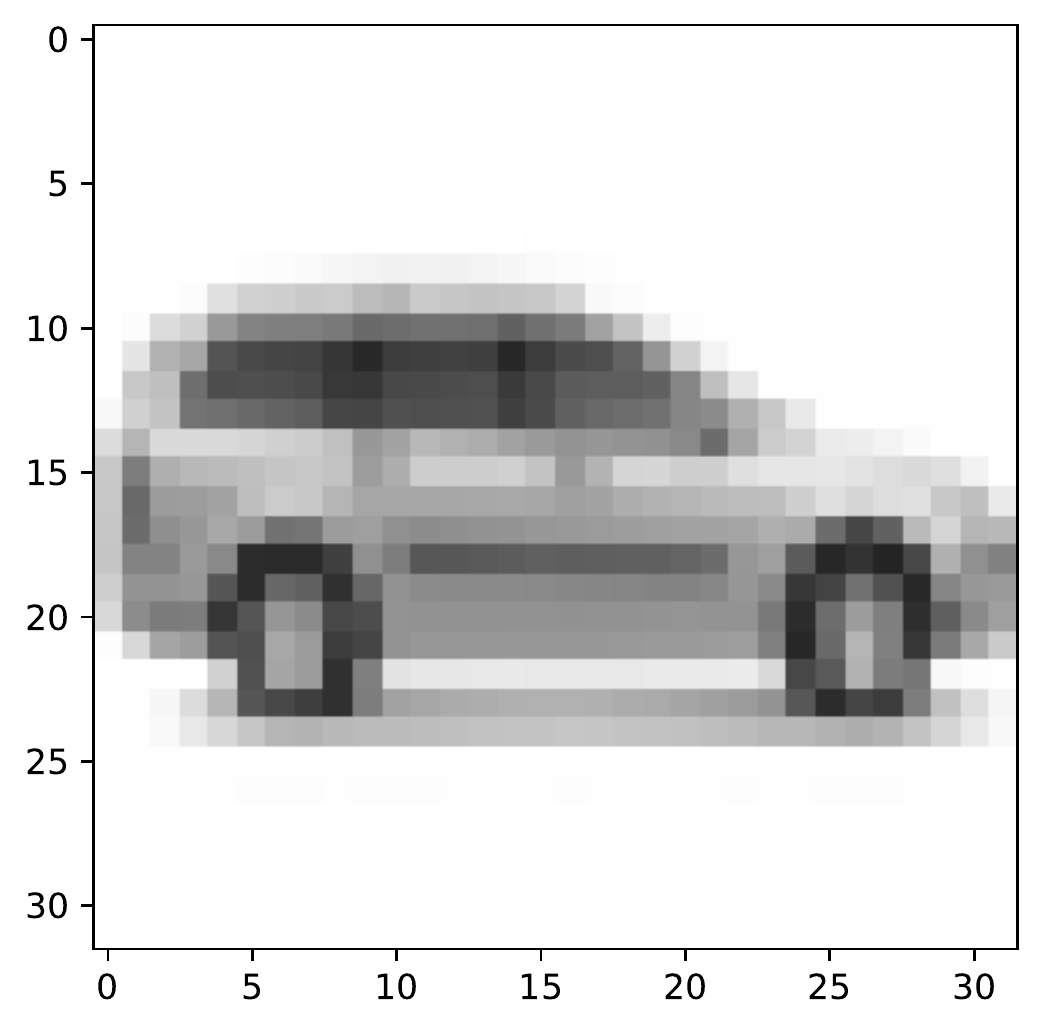} \\
    \end{tabular}
    \caption{a) Original (undistorted) gray-scale image, b) input image corrupted by Gaussian noise with standard variation $100$, images retrieved by c) bipolar ECNN, d) complex-valued multistate ECNN with $K=246$, e) quaternion-valued multistate ECNN with $K=16$, and split-sign octonion-valued ECNN using asynchronous update.}
    \label{fig:CIFAR_Images}
\end{figure}

The $P=200$ hypercomplex-valued vectors have been stored in bipolar, complex-valued, quaternion-valued, and octonion-valued ECNN models. The parameters used in this experiment can be found in Table \ref{tab:parameters}. Furthermore, we introduced Gaussian noise into one of the selected images and, in a similar fashion, we converted the corrupted image into a hypercomplex-valued vector which has been fed as input to the hypercomplex-valued ECNNs. Figure \ref{fig:CIFAR_Images} depicts a gray-scale image from the CIFAR dataset, its corresponding version corrupted by Gaussian noise with standard variation $100$, and the images retrieved by the hypercomplex-valued ECNNs using asynchronous update. Note that the complex-valued and quaternion-valued multistate ECNN succeeded in the retrieval of the original horse image. In contrast, the bipolar and octonion-valued models produced a different image as output; they failed to retrieve the undistorted horse image. Recall that the jupyter notebook of this experiment, implemented in \texttt{Julia language}, can be found in \url{https://github.com/mevalle/Hypercomplex-Valued-Recurrent-Correlation-Neural-Networks}. 

\begin{figure}[t]
    \centering
    \includegraphics[width=1\columnwidth]{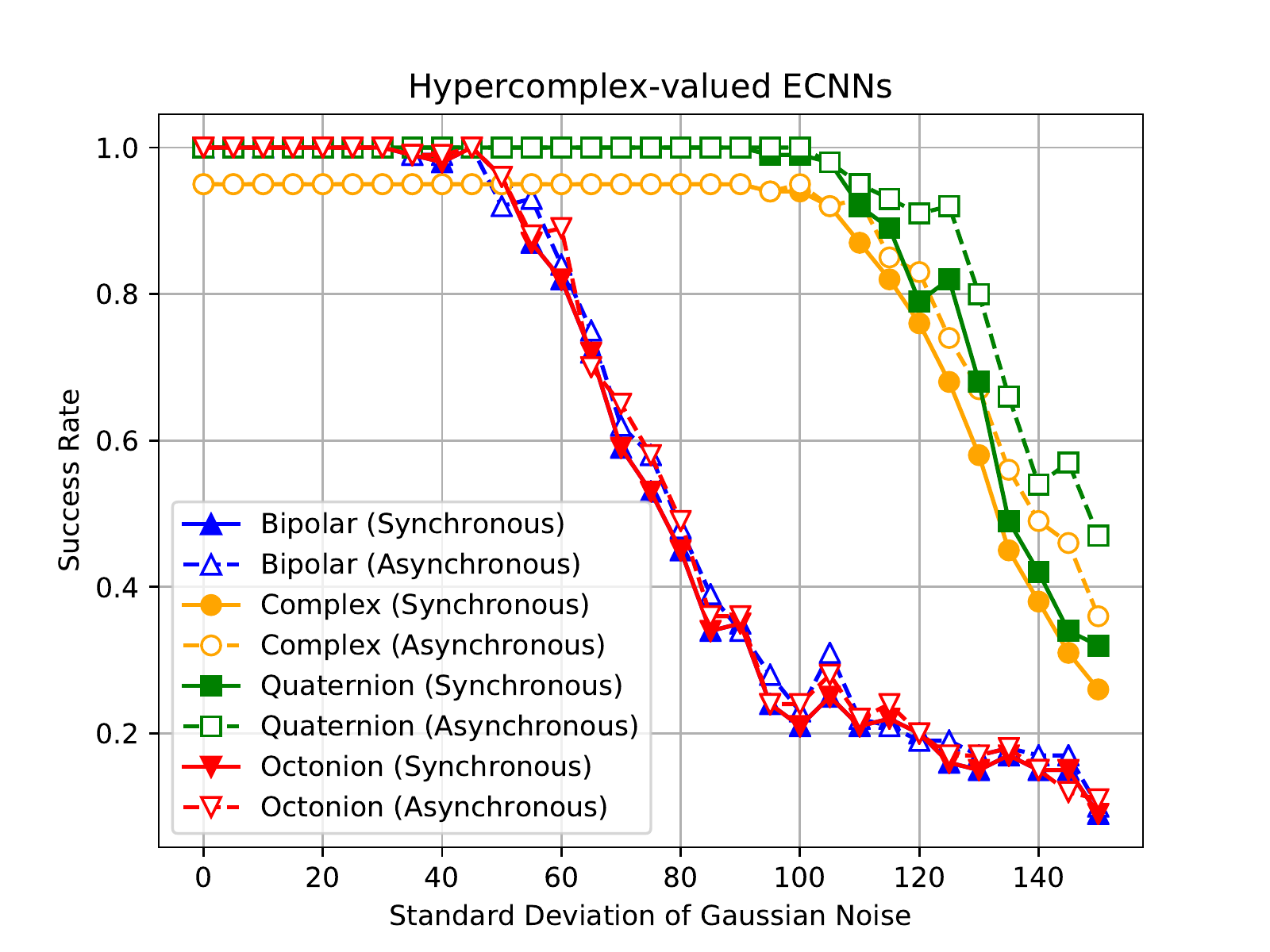}
    \caption{Probability of successful recall by the standard variation of Gaussian noise introduced into the input image.}
    \label{fig:CIFAR_ALL}
\end{figure}
Let us now compare the noise tolerance of the four synchronous hypercomplex-valued ECNNs designed for the storage and recall of images from the CIFAR dataset. To this end, we randomly selected $P=200$ images and corrupted one of them with Gaussian noise with several different standard variation. We repeated the experiment 100 times and computed the probability of successful recall of the original image. Figure \ref{fig:CIFAR_ALL} depicts the outcome of this experiment using both synchronous and asynchronous update.  Note that the quaternion-valued model yielded the largest noise tolerance among the four hypercomplex-valued ECNNs. Furthermore, the asynchrounous update yielded a probability of successful recall larger than the synchronous update. The second largest noise tolerance was produced by the complex-valued ECNN model using asynchronous update. Finally, the bipolar and the octonion-valued ECNN coincide using the synchronous update. Also, there is no significant difference between the bipolar and octonion-valued models synchronous and asynchronous update modes. Concluding, this experiment reveals the potential application of the new hypercomplex-valued ECNNs as associative memories designed form the storage and recall of gray-scale images. 

\section{Concluding Remarks} \label{sec:concluding}

Hypercomplex-valued neural networks constitute an emerging research area. Besides their natural capability to treat high-dimensional data as single entities, some hypercomplex-valued neural networks can cope with phase or rotation information \cite{aizenberg11book,hirose12,parcollet19air} while others exhibit fast learning rules \cite{nitta18}. 

In this paper, we introduced a broad class of hypercomplex-valued recurrent correlation neural networks (RCNNs). RCNNs have been introduced by Chiueh and Goodman in the early 1990s as an improved version of the famous correlation-based Hopfield neural network \cite{chiueh91}. In fact, some RCNNs can reach the storage capacity of an ideal associative memory \cite{hassoun96}. Furthermore, they are closely related to the dense associative memories of Krotov and Hopfield \cite{krotov16} as well as support vector machines and the kernel trick \cite{garcia04a,garcia04b,perfetti08}. In particular, the exponential correlation neural network (ECNN), besides been amiable for very-large-scale integration (VLSI) implementation, has a Bayesian interpretation \cite{chiueh93,hancock98}. 

Precisely, in this paper, we first reviewed the basic concepts on hypercomplex-number systems. Briefly, the key notions for the stability analysis of hypercomplex-valued RCNNs is the class of $\mathcal{B}$-functions (see Definition \ref{def:B-projection}) \cite{castro20nn}. Indeed, Theorem \ref{thm:Convergence} shows that hypercomplex-valued RCNNs always settle at an equilibrium, using either synchronous or asynchronous update modes, if the activation function in the output layer is a $\mathcal{B}$-function. Examples of the bipolar, complex, hyperbolic, quaternion, and octonion-valued RCNNs are given in Section \ref{sec:examples} to illustrate the theoretical results. Furthermore, Section \ref{sec:examples} presents a detailed experiment concerning the storage and recall of natural gray-scale images.  The quaternion-valued ECNN with the twin-multistate activation function exhibited the largest noise tolerance in this experiment. Also, the asynchronous update yielded a probability of successful recall larger than the synchronous update. 

In the future, we plan to study hypercomplex-valued RCNN defined on different hypercomplex-number systems as well as models with alternative activation functions. We also intend to investigate further applications of the new models, for example, in classification problems. Finally, based on our recent paper \cite{valle19bracis}, we plan to develop a broad class of hypercomplex-valued projection-based neural networks.

\appendix
\section{Some Remarks on the Computational Implementation of Hy\-per\-complex-valued Exponential Correlation Neural Networks}
\label{appendix}

In our codes, a hypercomplex-valued vector $\vetx(t) = [x_1(t),\ldots,x_N(t)]$, where $x_i(t) = \hyper{{x_i}}{n}$ for all $i=1,\ldots,N$, is represented by a $(N,n+1)$-array of real numbers:
\[ \mathtt{x} = \begin{bmatrix} 
x_{10} & x_{11} & \ldots & x_{1n} \\
x_{20} & x_{21} & \ldots & x_{2n} \\
\vdots & \vdots & \ddots & \vdots \\
x_{N0} & x_{N1} & \ldots & x_{Nn}
\end{bmatrix}. \]
Similarly, the fundamental memory set $\mathcal{U}=\{\vetu^1,\ldots,\vetu^P\}$ is represented by a $(N,n+1,P)$-array $\mathtt{U}$ such that $\mathtt{U(:,:,\xi)}$ corresponds to $\vetu^\xi$, for $\xi=1,\ldots,P$.

An arbitrary hypercomplex-valued ECNN is implemented as a function which receives the following inputs and outputs the final vector state and an array with the evolution of the function $E$ given by \eqref{eq:energy}:
\begin{enumerate}
    \item The symmetric bilinear form $\mathcal{B}$ and its parameters.
    \item The activation function $\phi$ and its parameters.
    \item The array $\mathtt{U}$ corresponding to the fundamental memory set.
    \item The array $\mathtt{x}$ corresponding to the input vector $\vetx(0)$.
    \item The optional parameters $\alpha$ and $\beta$ as well as the maximum number of iterations.
\end{enumerate}
Note that both hypercomplex-valued product, as well as the reverse-involution, are implicitly defined in the symmetric bilinear form $\mathcal{B}$ which is passed as an argument to the hypercomplex-valued ECNN model. 

Finally, we would like to point out that there is one code for the synchronous hypercomplex-valued ECNN model and another code for the asynchronous version. To fasten the implementation of the asynchronous hypercomplex-valued ECNN, instead of using \eqref{eq:weights}, we updated the weights $w_\xi(t+1)$ as follows for $\xi=1,\ldots,P$: Suppose only the $i$th neuron have been updated at time $t$, that is, $x_j(t+1)=x_j(t)$ for all $j \neq i$. Then, the weight $w_\xi(t+1)$ satisfies
\bb \label{eq:weight_update}
    w_\xi(t+1) 
    = w_\xi(t) \exp \left\{\alpha \mathcal{B}\big( u_i^\xi, x_i(t+1) - x_i(t) \big) \right\}.
\ee 
Note that, in contrast to \eqref{eq:weights} which evaluates $N$-times the symmetric bilinear form $\mathcal{B}$, \eqref{eq:weight_update} requires a single evaluation of $\mathcal{B}$. 




\bibliographystyle{elsarticle-num-names}
\bibliography{mail.bbl}







\end{document}